\documentclass[sn-mathphys-num,iicol]{sn-jnl}

\usepackage{graphicx}%
\usepackage{multirow}%
\usepackage{amsmath,amssymb,amsfonts,mathtools}%
\usepackage{amsthm}%
\usepackage{mathrsfs}%
\usepackage[title]{appendix}%
\usepackage{xcolor}%
\newif\ifshowchanges
\definecolor{revNineColor}{RGB}{0,128,64}
\definecolor{revTenColor}{RGB}{200,0,0}
\definecolor{revBothColor}{RGB}{128,0,160}
\definecolor{revLangColor}{RGB}{225,110,0}
\DeclareRobustCommand{\revNine}[1]{{\color{revNineColor}#1}}
\DeclareRobustCommand{\revTen}[1]{{\color{revTenColor}#1}}
\DeclareRobustCommand{\revBoth}[1]{{\color{revBothColor}#1}}
\DeclareRobustCommand{\revLang}[1]{{\color{revLangColor}#1}}
\ifshowchanges\else
\definecolor{revNineColor}{gray}{0}\definecolor{revTenColor}{gray}{0}\definecolor{revBothColor}{gray}{0}\definecolor{revLangColor}{gray}{0}
\DeclareRobustCommand{\revNine}[1]{#1}\DeclareRobustCommand{\revTen}[1]{#1}\DeclareRobustCommand{\revBoth}[1]{#1}\DeclareRobustCommand{\revLang}[1]{#1}
\fi
\AtBeginDocument{\pdfstringdefDisableCommands{\def\revNine#1{#1}\def\revTen#1{#1}\def\revBoth#1{#1}\def\revLang#1{#1}}}
\usepackage{textcomp}%
\usepackage{manyfoot}%
\usepackage{booktabs}%
\usepackage{algorithm}%
\usepackage{algorithmicx}%
\usepackage{algpseudocode}%
\usepackage{listings}%
\usepackage{relsize}
\usepackage{array,threeparttable,caption}
\usepackage{siunitx}
\usepackage{makecell}
\usepackage{tabularx}
\usepackage{booktabs,tabularx}

\usepackage{orcidlink}

\theoremstyle{thmstyleone}%
\theoremstyle{thmstyletwo}%

\theoremstyle{thmstylethree}%

\begin{document}

\title[Real-Time Remote Photoplethysmography (rPPG)]{Design, Implementation and Evaluation of a Real-Time Remote Photoplethysmography (rPPG) Acquisition System for Non-Invasive Vital Sign Monitoring}



\author*[1]{\fnm{Constantino} \sur{Álvarez Casado}\orcidlink{0000-0002-3052-4759}}\email{constantino.alvarezcasado@oulu.fi}

\author[1]{\fnm{Sasan} \sur{Sharifipour}\orcidlink{0000-0001-6670-4252}}\email{sasan.sharifipour@oulu.fi}
\author[1]{\fnm{Manuel} \sur{Lage Cañellas}\orcidlink{0000-0002-4917-340X}}\email{manuel.lage@oulu.fi}
\author[1]{\fnm{Nhi} \sur{Nguyen}\orcidlink{0009-0002-2090-0746}}\email{thi.tn.nguyen@oulu.fi}
\author[1]{\fnm{Le} \sur{Nguyen}\orcidlink{0000-0001-7765-1483}}\email{le.nguyen@oulu.fi}

\author[1,2]{\fnm{Miguel} \sur{Bordallo López}\orcidlink{0000-0002-5707-9085}}\email{miguel.bordallo@oulu.fi}

\affil*[1]{\orgdiv{Center for Machine Vision and Signal Analysis}, \orgname{University of Oulu}, \orgaddress{\city{Oulu}, \country{Finland}}}


\affil[2]{\orgname{VTT Technical Research Centre of Finland}, \orgaddress{\city{Oulu}, \country{Finland}}}


\abstract{The growing integration of smart environments and low-power computing devices, coupled with mass-market sensor technologies, is driving advancements in remote and non-contact-based physiological monitoring. However, deploying these systems in real-time on resource-constrained platforms introduces significant challenges related to scalability, interoperability, and performance.
This paper presents a \revTen{computationally efficient} real-time remote photoplethysmography (rPPG) system, designed to extract physiological signals, such as heart rate (HR), respiratory rate (RR), and oxygen saturation (SpO2), from facial video streams. The system is built on the Face2PPG pipeline, which processes video frames sequentially for rPPG signal extraction and analysis, while leveraging a multithreaded architecture to manage video capture, real-time processing, network communication, and graphical user interface (GUI) updates concurrently.
This design \revTen{targets continuous operation} at 30 frames per second (FPS), with adaptive feedback through a collaborative user interface to guide optimal signal capture conditions. The network interface includes both an HTTP server for continuous video streaming and a RESTful API for on-demand vital sign retrieval.
To ensure accurate performance despite \revTen{limited computational resources}, we use a hybrid programming model combining Functional Reactive Programming (FRP) and the Actor Model, allowing event-driven processing and efficient task parallelization.
\revBoth{We evaluate HR accuracy on three public datasets (COHFACE, LGI-PPGI, and UBFC-RPPG), and conduct preliminary evaluations of RR on COHFACE and SpO$_2$ on PURE. We also assess the processing time and end-to-end operation of the system on two computers, including live HR and SpO$_2$ agreement with a medical-grade fingertip pulse oximeter across five 10-minute sessions.} Our work addresses key challenges in real-time \revBoth{contactless} biosignal monitoring. \revTen{Code available at: \url{https://github.com/Multimodal-Sensing-Lab/rt-rppg}.}}





\maketitle

\section{Introduction}\label{sec:introduction}

The ongoing digital transformation in healthcare and well-being is reshaping care pathways \cite{taylor2023clinical}\cite{dunn2024remote}, introducing new challenges and opportunities, particularly in remote healthcare monitoring and telemedicine \cite{huang2023challenges}\cite{albahri2018systematic}. This transformation is driven by the need for efficient, scalable, and non-invasive methods to monitor patient health outside traditional clinical settings \cite{malasinghe2019remote}. Telemedicine, which often involves appointments conducted via videocalls, has become increasingly prevalent \cite{huang2023challenges}\cite{malasinghe2019remote}. However, these interactions largely depend on self-reported data from patients and the occasional use of home medical devices, such as pulse oximeters or blood pressure monitors, which patients must know how to operate. This approach is not only limited but also inconsistent in its application and accuracy.

With the growing need for continuous and unobtrusive health monitoring, contactless technologies using cameras or millimeter wave radars (mmWave) are emerging as a promising solution \cite{Face2PPGPipeline2022}\cite{antolinos2020cardiopulmonary}. These technologies leverage techniques such as remote photoplethysmography (rPPG) and remote ballistography (rBSG) to measure physiological parameters without direct contact \cite{alvarez2023biosignal}. rPPG, in particular, uses optical sensors and ambient light to detect changes in blood volume in the microvascular bed of tissue, allowing the extraction of vital signs such as heart rate (HR), oxygen saturation (SpO2) and respiratory rate (RR). Studies have demonstrated a high agreement between rPPG measurements and traditional medical devices under certain conditions, making it a promising method for non-invasive monitoring \cite{dasari2021evaluation}. Current methods for recovering physiological signals from videos are mainly divided into unsupervised non-learning-based methods and deep learning-based approaches \cite{Face2PPGPipeline2022}. The former use carefully-designed computer vision and signal processing algorithms\revLang{,} while the latter provide end-to-end solutions\revLang{,} yet they heavily rely on training data.

While rPPG methods \revLang{have high potential}, there is a significant gap between current research and practical deployment in real world systems \cite{albahri2018systematic}. Existing studies primarily focus on maximizing accuracy using well-curated, publicly available datasets, without addressing the real-time implementation challenges \cite{Face2PPGPipeline2022}\cite{Nguyen2024EvaluationOV}. These challenges include computational and resource constraints of low-power embedded systems, as well as communication issues \cite{Nguyen2024EvaluationOV}. To bridge this gap, the algorithms need to be optimized for low power consumption, efficient memory usage, and robustness against frame loss. This involves proper scheduling, synchronization, and strategies to handle frame loss and artifacts such as jitter caused by network quality, ensuring the applicability of the system in realistic scenarios \cite{albahri2018systematic}\cite{Nguyen2024EvaluationOV}.

\textcolor{black}{To meet the demands of next-generation telemedicine \revBoth{\cite{drees2020topol}} and \revBoth{patient-facing AI systems \cite{topol2026future}}, there is a growing need for edge-capable, modular, and transparent systems that support real-time physiological monitoring in unconstrained environments \cite{ackerman2010developing}. These systems not only improve usability and accessibility in clinical and home settings, but also form the foundation for emerging hybrid intelligence paradigms, in which artificial systems interpret human physiological and emotional states to support informed, empathetic decision-making \cite{jiang2024human}. Real-time implementations also serve a critical exploratory role, revealing system-level challenges such as frame drops, latency, and hardware bottlenecks that often go unnoticed in offline evaluation. Addressing these factors during system development can inform more robust algorithm design, encourage practical trade-offs between complexity and responsiveness, and ultimately enable more trustworthy and scalable deployment \cite{buttazzo2006research}. Therefore, designing and validating real-time rPPG systems for low-power platforms is not only a step toward technical feasibility, but a prerequisite for integrating physiological sensing into everyday intelligent systems.}

%
%
In this context, this article proposes the design and implementation of a real-time rPPG system that aims to measure heart rate and other cardiovascular indicators.
This system is \revTen{designed for computational efficiency on resource-constrained platforms}, incorporating strategies to mitigate artifacts resulting from network and hardware limitations.
Our approach includes a multithreaded architecture to handle image capture, processing, and data communication, \revTen{targeting} stable operation at 30 frames per second (FPS).
\revTen{By distributing the processing tasks across multiple CPU threads}, we achieve significant performance improvements, enabling the system to function effectively even under constrained conditions. The main contributions are detailed as follows:

\begin{itemize}
\item Developing a prototype for extracting rPPG signals from human faces in real-time, using a pipeline architecture for continuous, efficient vital sign monitoring.
\item Optimizing the rPPG system for \revTen{resource-constrained devices} with algorithms designed to minimize \revTen{computational cost} and memory usage, ensuring robust real-time performance.
\item \revBoth{Conducting speed and heart rate accuracy assessments using three public datasets, demonstrating the effectiveness of the system under different recording conditions for remote health monitoring.}
\end{itemize}

\revBoth{Although the system estimates HR, RR and SpO2, the quantitative evaluation presented in this work focuses on HR, with preliminary evaluations of RR and SpO2 on one dataset each. Most public rPPG datasets provide reference PPG or HR signals, whereas suitable references for RR and, especially, SpO2 are scarce.}

This work addresses the practical challenges of deploying rPPG technology in realistic scenarios, providing a scalable and efficient solution for continuous, non-invasive health monitoring. Our findings offer valuable insights for future digital healthcare systems, supporting the broader adoption of telemedicine and remote monitoring technologies.

%
%
%
%

\section{Related work}
\label{sec:1}
The implementation of real-time rPPG systems poses multiple challenges, especially when deployed on low-power embedded devices. These challenges can be broadly categorized into intrinsic issues related to the technology itself and extrinsic factors that arise due to the operational environment conditions \cite{Nguyen2024EvaluationOV}.

\subsection{\revNine{Intrinsic challenges in rPPG systems}}
Camera-based systems, such as those used in rPPG systems, often face challenges related to variations in motion and noise \cite{Face2PPGPipeline2022}, ambient lighting changes and low-light conditions \cite{Lin2017IlluminationColor}, occlusions \cite{Nguyen2024EvaluationOV}, camera distance and resolution \cite{song2020DistanceFactor}, and skin tone \cite{Setchfield2024TypeSkin}\cite{kumar2015distanceppg}. These challenges are particularly pronounced when these systems are deployed in real-world settings, beyond the controlled conditions of a laboratory, and can severely affect the accuracy and quality of the extracted biosignals \cite{Face2PPGPipeline2022}\cite{Nguyen2024EvaluationOV}. These variables can introduce noise and artifacts into the captured data, complicating signal processing and reducing the accuracy of DL-based solutions, making reliable vital sign measurement more difficult. Nguyen et al. (2024) emphasize that effective handling of such artifacts is crucial to improve the resilience of rPPG systems under diverse operational conditions \cite{Nguyen2024EvaluationOV}.

\subsection{\revNine{Extrinsic challenges affecting performance}}
Beyond the intrinsic challenges of rPPG methodologies and video signal extraction, various external factors can significantly impact the performance of real-time rPPG systems in practical settings. While these systems may perform optimally in controlled laboratory environments, real-world conditions introduce complexities that can compromise their effectiveness. Factors such as sensor limitations \cite{Nguyen2024EvaluationOV}, network artifacts and limitations \cite{Narayanan2021Networks5G}, and the processing capabilities of embedded devices \cite{kim2021RTVideoSurvillance} are critical during deployment. However, their influence is often underestimated in research. Few studies have thoroughly examined these aspects, which are crucial in designing and implementing real-time rPPG methods. This oversight is particularly significant in video streaming for real-time vital sign monitoring, where maintaining signal quality with low latency and limited bandwidth is essential. Despite advancements in 5G and 6G technologies, suboptimal network conditions can still occur, impacting rPPG performance. Techniques such as robust collaborative-relay beamforming can help mitigate video artifacts and network constraints, ensuring reliable real-time vital sign monitoring on wireless devices like smartphones and laptops \cite{zhao2018optimization}. Nevertheless, external factors such as network and hardware limitations must be accounted for, as they can significantly affect the performance of rPPG systems. Issues such as video codecs and video compression \cite{McDuff2017VideoCodecs}, video resolution \cite{Nguyen2024EvaluationOV}\cite{song2020DistanceFactor}, frame rate variations \cite{Nguyen2024EvaluationOV}\cite{Blackford2015FrameRateReso}, and network jittering and frame loss due to network instability \cite{Nguyen2024EvaluationOV} can disrupt the continuity and integrity of data, essential for smooth biosignals extraction and analysis. Video compression can introduce noise and spatio-temporal artifacts into the biosignals \cite{hanfland2016videocomp}\cite{Zhao2018rPPGinCompressed}. Similarly, variations in frame rate and resolution can alter the morphology and frequency features of the extracted signals, impacting the accuracy of physiological signal extraction \cite{Nguyen2024EvaluationOV}. Research has also shown the importance of choosing the best image format to extract biosignals with better SNR and quality \cite{botina2022rtrppg}.

\subsection{\revNine{Real-time rPPG methods and systems}}
These factors should be considered during the design phase to avoid unexpected performance issues in real-world applications. Optimizing systems that perform well under lab conditions for real-world deployment often requires compromises, affecting expected performance. Addressing the above challenges requires not only robust algorithm design but also novel system architectures that can accommodate the demands of real-time processing. Botina-Monsalve et al. (2022) present RTrPPG, an ultra-light 3DCNN model designed to balance accuracy and processing time for real-time applications \cite{botina2022rtrppg}. This model optimizes the architecture by reducing input image dimensions and using a temporal-frequency-based loss function. These adjustments significantly improve inference speed, achieving 28.6 milliseconds per frame on an Intel Xeon CPU at 2.4 GHz and 2.32 milliseconds on an Nvidia GeForce RTX 2070 GPU. While these improvements make RTrPPG suitable for real-time deployment, it may still face challenges in low-powered embedded systems. Hasan et al. (2022) adapted the  deep learning models for resource-constrained
devices leveraging pruning and quantization techniques~\cite{Hasan2022}. Their approach reduced the model size, power consumption, memory usage, and latency to deploy the rPPG systems on edge platforms, including NVIDIA Jetson Nano, Google Coral Development Board, and Raspberry Pi.

\revNine{Other works have presented complete systems for real-time physiological monitoring with cameras. Gudi et al. (2020) proposed an unsupervised rPPG pipeline that estimates HR and heart rate variability from RGB webcams, combining active appearance modelling of the face, motion suppression based on the head pose and a wide and narrow band filtering of the pulse signal \cite{Gudi2020RealTime}. The complete analysis runs in real time on a CPU, at approximately 30 FPS for 640x480 videos on an Intel Xeon E5 processor, and the authors also introduced the VicarPPG 2 dataset and a collection of human-verified heartbeat annotations for existing rPPG datasets. Pirzada et al. (2022) presented ARPOS, an automated remote pulse oximetry system that estimates HR and SpO2 from the color, infrared and depth streams of a Microsoft Kinect V2 camera \cite{Pirzada2022ARPOS}. The system was evaluated with 40 participants recorded in their homes, using a commercial fingertip pulse oximeter as reference, and the HR and SpO2 estimates of the evaluation were obtained by processing the recorded face data offline.}


\subsection{\revNine{Hardware optimizations for embedded devices}}
\revNine{The} development of specialized hardware accelerators and optimized neural network architectures contributes to enhancing the feasibility of deploying advanced computer vision and signal processing techniques on low-power devices \cite{reuther2021ai}. These accelerators, including AI-specific units, enhance performance and efficiency, making advanced techniques more feasible in constrained environments. Examples include the Sagitta accelerator \cite{zhou2023sagitta}, which is designed for specific deep learning architectures and improves energy efficiency and reduces latency. AI accelerators like the Neural Processing Unit (NPU) in the Qualcomm Snapdragon 855, the Intel Myriad X Visual Processing Unit (VPU), and the Nvidia Jetson Nano with its Maxwell architecture offer substantial computational power beyond standard GPUs \cite{li2020survey}, especially for DL-powered systems. Open-source frameworks such as OpenVINO, ExecuTorch from PyTorch, ONNX Runtime (ORT), and AMD RyzenAI optimize and accelerate model inference on these specialized hardware platforms. However, not all end-users have access to these accelerators, and integrating them can be complex and costly \cite{li2020survey}. While beneficial for DL models, these accelerators offer limited advantages for traditional computer vision or signal processing tasks. Reliance on specific hardware can also limit portability and flexibility across different platforms. General-purpose frameworks like OpenCL, OpenMP, OpenACC, or OpenGL leverage the computing power of available hardware in embedded devices. These frameworks provide a versatile environment for parallel processing tasks \cite{czarnul2020survey}, improving performance for a wide range of applications. The advantage of using OpenCL and OpenGL is their broad compatibility with various hardware, which enhances portability \cite{czarnul2020survey}. However, these frameworks may not offer the same level of optimization for DL-based tasks as specialized AI accelerators. Considering these factors during design ensures practical and effective solutions for diverse deployment scenarios.


\subsection{\revNine{Scheduling and multithreading}}
Effective real-time systems require well-structured software architectures and precise concurrency control mechanisms \cite{Kopetz2022}. To achieve optimal performance and responsiveness, researchers and practitioners have explored the potential of multithreading and multiprocessing \cite{salehian2017comparison,buono2014optimizations,Berjon2011Parallel}.
Buttazzo (\revNine{2023}) emphasizes the critical role of predictable scheduling algorithms in ensuring timely task execution within constrained resource environments \cite{Buttazzo2024}. These algorithms can be implemented at multiple levels, including hardware, low-level software, and the application layer \cite{buono2014optimizations}. Predictable scheduling can be implemented within the application itself, especially in real-time stream-processing applications that involve multiple subtasks needing coordination. This approach is essential for ensuring tasks execute within strict time constraints, managing task assignment to system resources like CPU cores to guarantee completion within predictable limits \cite{Buttazzo2024}. Key features of predictable scheduling include determinism, which ensures that system behavior can be precisely predicted, timeliness, which guarantees that tasks meet their deadlines to avoid critical failures, and efficiency, which optimizes resource use to prevent idleness and enhance throughput. Lorenzon et al. (2015) highlight the benefits of multithreading in improving system performance, particularly on devices with limited computational resources, such as embedded systems \cite{lorenzon2015optimized}. Task parallelism involves breaking down complex tasks into smaller, independent subtasks that run concurrently, improving system throughput by using multiple cores \cite{vanneschi2014high,buono2014optimizations}. It includes decomposition (dividing tasks into subtasks), concurrency (executing subtasks independently), and asynchronicity (tasks proceed without waiting for others). Other important models include data parallelism, which distributes data across processors, thread-level parallelism, which uses multiple threads within a core, fork-join parallelism, which splits tasks into parallel threads and rejoins them, and hybrid parallelism, which combines task and data parallelism \cite{buono2014optimizations}. To coordinate concurrent activities effectively, event-driven synchronization at the application level is crucial \cite{leidi2011event}. This involves a main thread managing an event loop that interacts with other threads, handling task dependencies, and preventing performance bottlenecks. This method enhances system responsiveness by avoiding blocking operations and ensuring efficient resource use. Synchronization primitives like mutexes are essential for protecting shared resources from concurrent access. By enforcing mutual exclusion, mutexes prevent race conditions and data corruption. However, blocking mutexes can cause performance issues due to thread waiting. Non-blocking alternatives, such as spinlocks or optimistic concurrency control, offer more efficient solutions in specific scenarios \cite{lorenzon2015optimized}. The evolution of programming languages has significantly impacted the development of concurrent applications. Traditionally, low-level synchronization primitives, such as semaphores and condition variables, were primarily available through operating system APIs. However, modern languages increasingly incorporate higher-level concurrency constructs, such as promises/futures (e.g., C++11, Python, ECMAScript 7), monitors (e.g., Ada 95, Java), and parallel programming models like fork-join (e.g., OpenMP) and work stealing (e.g., Intel’s TBB, Apple’s GCD). These advancements simplify concurrent programming and enable developers to harness the power of multi-core processors more effectively \cite{salehian2017comparison,diez2016parallel}.

\subsection{\revNine{Positioning of the proposed system}}
\revNine{The real-time rPPG methods and systems described above focus mainly on the estimation methods, their evaluation, and the datasets. In contrast, our work focuses on the design and implementation of the real-time application itself, including its multithreaded architecture for video capture, processing, user feedback and network communication, and reports the processing time of each module of the pipeline.} Building on existing research, we have designed a real-time rPPG system based on a modular pipeline architecture, allowing easier optimization and parallelization. We optimized each pipeline block at both the algorithm and data levels. This included selecting algorithms with reduced computational complexity, implementing approximate methods where full precision was unnecessary, and choosing algorithms amenable to parallel processing. We incorporated pruning strategies to limit unnecessary computations and applied memoization by caching intermediate results to accelerate subsequent operations. Memory management was improved through the use of memory pooling and precise data alignment, which kept shared data structures persistent throughout execution, thus minimizing the overhead from frequent memory allocation and deallocation. Furthermore, we utilized data parallelism to effectively distribute processing tasks across multiple CPU cores, thereby improving overall performance and reducing latency. Additionally, we integrated real-time quality metrics to evaluate the input data, specifically estimating factors such as face movement. This approach ensures that only high-quality data is processed, thereby improving efficiency and speed by avoiding unnecessary computations on low-quality inputs. We implemented multithreading and task-parallel techniques using C++14, employing threads and mutexes to prevent data races. Asynchronous threads were utilized to operate independently, keeping the main processing flow unblocked and enhancing system responsiveness. This setup ensures the camera thread always provides the latest frame, eliminating the need for intentional frame-dropping strategies and maintaining consistent real-time performance. A central master thread maintains a regular frame rate and orchestrates the system. The design leverages multi-core and heterogeneous architectures, allowing tasks to run on dedicated CPU cores or with GPU support, and supports distributed processing. An event-driven graphical user interface (GUI) efficiently uses hardware resources, with the master thread generating events at regular intervals for continuous data capture and processing. These strategies improve performance and responsiveness, 
\revTen{making the system computationally efficient for real-time rPPG applications}. The following sections detail our rPPG pipeline and real-time architecture design, highlighting implementation details and benefits.

%
%
%
%

\section{Remote photoplethysmography pipeline}
\label{sec:2}

To extract rPPG signals from face videos, we use the unsupervised, non-learning-based Face2PPG pipeline \cite{Face2PPGPipeline2022}.
It integrates computer vision and signal processing techniques to deliver precise and reliable rPPG signal extraction, making it suitable for various applications in health monitoring and biometric analysis.
The pipeline begins with face detection and alignment, using either DL models or traditional ML methods. Subsequently, skin segmentation isolates regions of interest (ROIs) in the facial region for signal extraction. Dynamic ROI selection is used to identify areas with optimal signal quality while discarding regions with occlusion, low contrast, or potentially poor signals. The pipeline then extracts raw rPPG signals from the selected ROIs. These signals undergo filtering to remove noise, enhance quality, and isolate the frequency bands of interest. Following this, the RGB signals are transformed into physiological signals (PPG signals), and spectral analysis is performed to derive physiological parameters. Post-processing steps compute vital signs such as heart rate (HR), respiratory rate (RR), blood oxygen saturation (SpO2), and heart rate variability (HRV).
\revTen{In the real-time system presented in this work, the post-processing computes HR, RR and SpO2, while HRV is not computed.} Additionally, the pipeline incorporates movement and facial expression stabilization through geometric normalization using facial landmarks. This stabilization maintains consistency in ROI selection despite facial movements and expressions, ensuring robust and accurate signal extraction across several state-of-the-art datasets \cite{Face2PPGPipeline2022}.

\textcolor{black}{The adaptation of the Face2PPG pipeline for real-time operation, as detailed in the subsequent subsections, followed a systematic and iterative refinement process. Each module was critically analyzed and optimized to balance computational efficiency, \revTen{essential for real-time operation on resource-constrained devices}, with the signal fidelity required for reliable physiological measurement. This approach was informed by the practical challenges inherent in live video processing and real-time constraints, as highlighted by research trends in embedded systems \cite{buttazzo2006research}.} 




%
%
\subsection{Adapting and optimizing the Face2PPG pipeline to real-time operation}
\label{subsec:optimize}

The proposed real-time prototype was developed using C++ (C++14 standard) due to its performance and ability to manage system complexity. We used several optimized libraries: \textbf{OpenCV \revTen{4.5.5}} for computer vision tasks \cite{bradski2000opencv}, \textbf{Dlib 19.23} for machine learning tasks \cite{king2009dlib}, and \textbf{Qt 5.13.0} for developing a cross-platform GUI \cite{lazar2018mastering}. These libraries were selected for their functionalities and performance benefits, essential for real-time image processing applications. The adaptation of the Face2PPG pipeline for real-time operation involves the optimization and refining of the following sequential stages and modules:

\begin{enumerate}
\setlength\itemsep{2pt}

\item \textbf{Face Detection and Tracking}: Our real-time system uses two face detectors, each with unique characteristics. The first, YuNet by Wu et al. \cite{Wu2023yunet}, is a CNN-based detector known for its compactness and speed, with a size under one megabyte. Using the MobileNet structure \cite{howard2017mobilenets}, YuNet has 85,000 parameters, achieving detection rates of 909 FPS for 224x224 (1.1 ms), 625 FPS for 320x320 (1.6 ms), and 105 FPS for \revLang{640x480} (9.5 ms) on an Intel i7-12700K CPU in C++ over 1,000 iterations \cite{Wu2023yunet}. YuNet is optimized for real-time applications on CPUs, making it suitable for mobile and embedded devices while maintaining robust performance under varying conditions. The second detector, Dlib \cite{dlib09}, uses a method based on Histogram of Oriented Gradients (HOGs) \cite{Dalal2005HOGDetector}. This approach is effective for CPU use, particularly in detecting frontal and slightly non-frontal faces, and remains reliable even under minor occlusion. To maintain consistent face tracking, faces are detected every three frames, optimizing computational resources and enhancing real-time performance. This method updates a face flag to indicate the presence or absence of a face in the focus area (center rectangle marked in the GUI), allowing for effective tracking even if the face is temporarily obscured or moves out of focus. We use the face detectors as a primary step for applying a face alignment model on the detected faces. By asking users to stay still within the set area, we reduce movement errors, leading to more accurate estimations of vital signs.

\item \textbf{Face Alignment}: In the Face2PPG pipeline, precise facial alignment is crucial for geometric face normalization and accurate skin segmentation, important for extracting reliable physiological signals. This alignment also supports accurate tracking of head movements and facial expression analysis, providing insights into patient well-being. To ensure real-time performance on low-power computing systems, we optimized face alignment models for accuracy, speed, and memory-efficiency. While Deep Alignment Network (DAN) models offer high accuracy, their computational demands make them unsuitable for such environments. Instead, we focused on optimizing and training two fast, non-deep-learning-based models: Ensemble of Regression Trees (ERT) and Local Binary Features (LBF) algorithms. These models deliver rapid and precise face alignment in real-time. Our optimized models significantly outperform existing implementations of these algorithms \cite{AlvarezBordallo2021FaceAlignment}. Using the ERT algorithm in Dlib, we achieved a speed of up to 530 FPS on a CPU, offering a faster and more memory-efficient solution while maintaining high accuracy, even in challenging scenarios \cite{AlvarezBordallo2021FaceAlignment}. One such challenge is the \revLang{`jittering'} effect in facial videos, where small fluctuations in landmarks between frames can affect skin segmentation. Fast head movements can further disrupt landmark detection in real-time applications. To address these issues, we enhanced the training process and incorporated temporal information to improve landmark tracking in videos. A key strategy was integrating domain-specific data into training, including images with challenging poses or low-light conditions, annotated using a teacher-student approach, improving model performance under complex conditions \cite{AlvarezBordallo2021FaceAlignment}.

\item \textbf{Skin Segmentation and Region of Interest Selection}: The approach uses a novel geometric segmentation method based on facial landmark points \cite{Face2PPGPipeline2022}. The proposed segmentation method normalizes the face in each frame by mapping triangles from the detected face to a reference shape. This generates a spatiotemporal matrix of normalized faces, ensuring consistent signal measurement across frames, regardless of pose or movement. Although some landmark variability exists across video frames, it is reduced compared to skin color segmentation methods \cite{AlvarezBordallo2021FaceAlignment}. We have also developed a deep-learning-based skin segmentation method using \revLang{a U-Net} architecture.  However, it suffers \revLang{from} similar problems, caused by the small number of annotated facial skin masks, resulting in underfitted models \cite{nanni2023standardized}. We instead use an Alpha-Beta filter to stabilize the landmarks in each frame and reduce small changes. Regarding ROI selection, we strategically target three specific facial areas: cheeks and forehead. These regions, characterized by a dense network of blood vessels, are optimal for extracting biosignals from the face as demonstrated in \cite{Face2PPGPipeline2022}. This process facilitates the extraction of raw RGB signals from these areas, preparing them for the next steps.

\item \textbf{RGB to BVP Transformation (rPPG)}: The real-time system uses the CIE-Lab color space for RGB to BVP transformation. CIE-Lab was selected over methods such as CHROM, OMIT, and POS due to its efficient computation \cite{Face2PPGPipeline2022} without temporal dependency, allowing for independent conversion of each frame. Additionally, CIE-Lab provides a wider dynamic range than the simplest method (the green channel approach \cite{Lab2016SNR}) and effectively isolates the chrominance channel, which is essential for detecting blood-related signals \cite{Face2PPGPipeline2022}. Rather than converting the entire frame to CIE-Lab format, the system computes the average RGB values within the ROI and converts only these averages to CIE-Lab. This approach reduces the computational cost by performing a single conversion per ROI rather than processing every pixel. The conversion process is detailed as follows:

\begin{enumerate}

\item \textbf{Step 1: RGB to XYZ.}
First, the average RGB values are normalized to the range [0, 1]:
\begin{equation}
\footnotesize
R_{norm} = \frac{R}{255}, \quad G_{norm} = \frac{G}{255}, \quad B_{norm} = \frac{B}{255}
\end{equation}
Gamma correction is applied:
\begin{equation}
\small
R_{linear} =
\begin{cases}
\left(\frac{R_{norm} + 0.055}{1.055}\right)^{2.4}, & \text{if } R_{norm} > \beta \\
\frac{R_{norm}}{12.92}, & \text{if } R_{norm} \leq \beta
\end{cases}
\end{equation}
where $\beta=0.04045$.

The same correction is applied to $G_{linear}$ and $B_{linear}$. The linearized RGB values are then converted to XYZ using the following matrix:
\begin{equation}
\begin{pmatrix} X \\ Y \\ Z \end{pmatrix} =
\begin{pmatrix} 0.4124 & 0.3576 & 0.1805 \\ 0.2126 & 0.7152 & 0.0722 \\ 0.0193 & 0.1192 & 0.9503 \end{pmatrix}
\begin{pmatrix} R_{linear} \\ G_{linear} \\ B_{linear} \end{pmatrix}
\end{equation}

\item \textbf{Step 2: XYZ to CIE-Lab.} The XYZ values are normalized relative to the reference white point (D65: $X_n = 95.047$, $Y_n = 100.000$, $Z_n = 108.883$):
\begin{equation}
X' = \frac{X}{X_n}, \quad Y' = \frac{Y}{Y_n}, \quad Z' = \frac{Z}{Z_n}
\end{equation}
Each normalized value is transformed as follows:
\begin{equation}
f(v) =
\begin{cases}
v^{1/3}, & \text{if } v > 0.008856 \\
7.787v + \frac{4}{29}, & \text{if } v \leq 0.008856
\end{cases}
\end{equation}
Finally, the CIE-Lab values are computed:
\begin{equation}
L^* = 116 \cdot f(Y') - 16
\end{equation}
\begin{equation}
a^* = 500 \cdot \left[f(X') - f(Y')\right]
\end{equation}

\begin{equation}
b^* = 200 \cdot \left[f(Y') - f(Z')\right]
\end{equation}

\end{enumerate}

\vspace{3mm}
This optimized conversion method minimizes computational overhead while effectively targeting the chrominance channel, which is important for accurately extracting blood-related signals.

\item \textbf{Buffering and Windowing}: The system maintains a buffer implemented with a C++ \textit{deque} that holds up to 360 samples of the a-channel values from the CIE-Lab color space, corresponding to a 12-second window
\revTen{at 30 FPS. Similar buffers hold the L* channel values used for RR, the average red and green values used for SpO2 \revBoth{and their band-pass filtered values}, and the timestamp of each sample}. This buffer supports the real-time computation of HR and RR by continuously updating with the most recent data. The buffer functions in a rolling manner, where each new sample is added to the back (\textit{push\_back()}) and the oldest sample is removed from the front (\textit{pop\_front()}), ensuring no data is lost and calculations are always based on the latest available data.
\revTen{The system computes the vital signs every 30 samples (every second at 30 fps) as new data arrives, and the first values are displayed within the first 90 samples (3 seconds at 30 fps).} After \revTen{the first eight updates (240 samples, i.e., 8 seconds at 30 fps)} of uninterrupted data, the system marks the signal as stable, indicating that the data is reliable for further processing and transmission.

\revTen{The RR and SpO2 values are also computed at every update, and they are transmitted with the same stability flag as the HR.}
\revTen{Before the signal is stable, the HR is computed from the raw a-channel buffer. Once the signal is stable, the system shifts to the band-pass filtered signal, using up to 299 filtered samples (about 10 seconds at 30 FPS), while the RR and SpO2 are computed from the raw buffers of 360 samples (12 seconds at 30 FPS).} The buffer size and window duration are configured to optimize real-time performance while minimizing computational overhead.
\revTen{The buffers hold a fixed number of samples rather than a fixed duration, and the video frames are not resampled. At each update, the frame rate is estimated from the timestamps of the buffered samples, and it is used to convert the frequency of the spectral peaks into beats or breaths per minute. Therefore, the duration covered by the buffers depends on the actual frame rate, for instance 18 seconds for 360 samples at 20 FPS. Frames without a detected face do not add samples to the buffers, and the buffers are reset after 60 consecutive frames without a face.} The stability flag ensures that only uninterrupted and reliable data is used for post-processing and biosignal transmission, enhancing the overall accuracy and robustness of the system.

\item \textbf{Dynamic Signal Filtering}: Signal filtering is essential for isolating frequency components and reducing noise in physiological signals for improved HR and RR measurements. The system uses a 61-tap Finite Impulse Response (FIR) filter with a linear phase design, which preserves the temporal structure of the signals and avoids phase distortion. The filter is tailored for specific physiological ranges.
For HR, a band-pass filter (BPF) is applied within a range of \revTen{0.7 to 2.5 Hz (42 to 150 bpm)}. For RR, a \revTen{9-tap} low-pass filter (LPF) is set at 0.7 Hz, covering up to 42 breaths per minute\revTen{, and it is applied to the respiratory signal displayed in the GUI, while the RR is estimated from the unfiltered signal, as described below. The filter coefficients are computed for a nominal sampling rate of 30 Hz.}

The 61 taps in the BPF offer a balance between frequency selectivity and computational cost. More taps would sharpen the filter response but increase the processing load, while fewer taps would reduce filtering performance.

The filter coefficients are calculated using the sinc function to design the band-pass filter:

\begin{equation}
m = n - \frac{N - 1}{2}, \quad \text{for each } n
\end{equation}

\begin{equation}
h[n] =
\begin{cases}
\frac{\phi - \lambda}{\pi}, & \text{if } m = 0, \\
\frac{\sin(m \cdot \phi) - \sin(m \cdot \lambda)}{\pi \cdot m}, & \text{if } m \neq 0.
\end{cases}
\vspace{3mm}
\end{equation}
where \( N \) is the total number of filter taps, \( n \) is the index of the filter tap ranging from \( 0 \) to \( N - 1 \), and \( m \) is the index relative to the center of the filter. \( \phi \) and \( \lambda \) are the angular frequencies corresponding to the upper and lower cutoff frequencies, respectively. The value of \( h[n] \) represents the filter coefficient for the \( n \)-th tap of the filter. 

By storing past values of the input signal in a shift register, the filter maintains continuity across consecutive data windows. This minimizes transient effects that can occur at the beginning of each new window, ensuring smoother and more consistent filtering. The fixed filter coefficients were selected to ensure efficient and reliable signal extraction, while the buffering of past data reduces artifacts without adding significant computational overhead.

\item \textbf{Frequency Analysis}: The system performs frequency analysis on heart and respiratory signals using the Discrete Fourier Transform (DFT).
\revTen{Each signal is first normalized to zero mean and unit variance and multiplied by a Hann window to reduce spectral leakage. Then, it is zero-padded to five times its length for HR and to nine times its length for RR. Zero-padding interpolates the spectrum, providing a denser sampling of the frequency axis, but it does not improve the frequency resolution, which is determined by the duration of the analyzed signal.} The Power Spectral Density (PSD) is computed from the DFT output by summing the squared magnitudes of the real and imaginary components of each frequency bin. To ensure accurate frequency analysis despite frame rate variations, the system recalculates the frame rate \revTen{at every update} based on captured timestamps.
The HR is determined from the dominant frequency in the PSD, within the range of \revTen{0.67 to 2.5 Hz (40 to 150 bpm)}, after filtering out noise outside this range. The system refines HR estimation iteratively by excluding erroneous peaks and outliers.
\revBoth{The RR is estimated from the L* channel values of the central facial region, as the power-weighted mean frequency of their spectrum within the range of 8 to 42 breaths per minute (0.13 to 0.70 Hz), where 8 breaths per minute is the lower frequency limit of the respiratory band.}
For increased robustness, Welch's method \revTen{was also evaluated as an alternative in the offline evaluation of one of the real-time configurations, for} cases where the signal may exhibit noise or instability due to motion artifacts or minor inconsistencies in the signal capture process. By dividing the signal into overlapping segments and averaging their periodograms, Welch's method reduces spectral variance, providing more stable HR and RR estimates in challenging scenarios.
Although this method requires additional computational resources, it \revTen{can improve} the reliability of the frequency analysis when the signal is less stable or contains transient disturbances.

\item \revBoth{\textbf{SpO2 Estimation}: The SpO2 is estimated with the ratio-of-ratios principle of pulse oximetry, adapted to the red and green channels of the camera \cite{vanGastel2022CalibratedSpO2}. The average red and green values of the central facial region are stored in the buffers described above, together with their values filtered with a band-pass filter of 0.75 to 1.75 Hz (a 61-tap FIR filter designed as described in item 6), which retains the cardiac pulsation. At every update, the pulsatile (AC) component of each channel is approximated by the standard deviation of its filtered values, and the non-pulsatile (DC) component by the mean of its unfiltered values. The SpO2 is then computed as:}
\begin{equation}
\label{eq:spo2}
\revBoth{\mathrm{SpO_2} = A - B \cdot \frac{\lambda_G \cdot AC_G / DC_G}{\lambda_R \cdot AC_R / DC_R}}
\end{equation}
\revBoth{where the subscripts $G$ and $R$ denote the green and red channels, $\lambda_G = 550$ and $\lambda_R = 850$ are the weights associated with each channel, and $A = 100$ and $B = 2$ are empirical constants. These constants were not calibrated against reference SpO2 measurements, so the SpO2 values provided by the current implementation are an uncalibrated approximation, and they are only evaluated in a preliminary way in this work.}

\end{enumerate}

%
%
%
%
\section{Real-time system architecture design and implementation}

\begin{figure*}[ht!]
\begin{center}
\includegraphics*[width=0.99\textwidth]{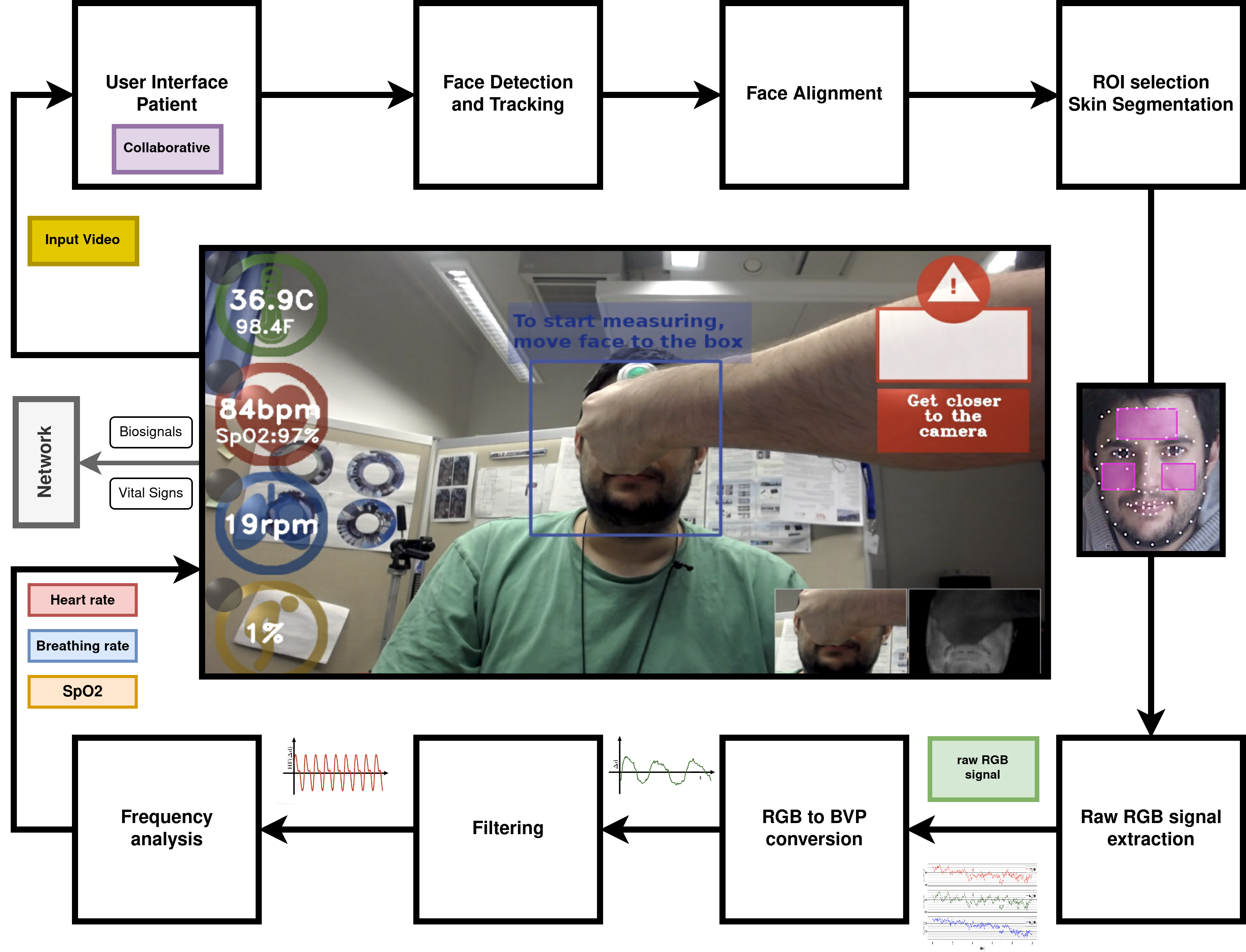}
\end{center}
\caption{Architecture overview of the proposed real-time rPPG application, comprised of several interconnected components, each executing a specific task.}
\label{fig:modular_arch_rt_app}
\end{figure*}

Building on the Face2PPG rPPG extraction pipeline and the remote biosignal acquisition framework described in \cite{alvarez2024distributed}, we developed a real-time application for contactless physiological monitoring using facial video streams. The application extracts rPPG signals and estimates vital signs, with potential applications in video consultations and remote healthcare assessments \cite{drees2020topol}. The core processing pipeline follows a modular architecture in which each stage processes the output of the preceding one, as illustrated in Figure \ref{fig:modular_arch_rt_app}. This organization separates the signal processing stages while allowing other system components, such as video capture, user interface updates, and network communication, to operate concurrently. 

The architecture was progressively refined through iterative implementation and testing. Practical challenges observed during real-time operation, including frame drops, interface responsiveness, and task synchronization, guided the design and scheduling decisions. This process helped balance computational efficiency, system responsiveness, and the requirements of continuous physiological signal acquisition \cite{de2020modular}.

\begin{figure}[ht!]
\begin{center}
\includegraphics*[width=0.49\textwidth]{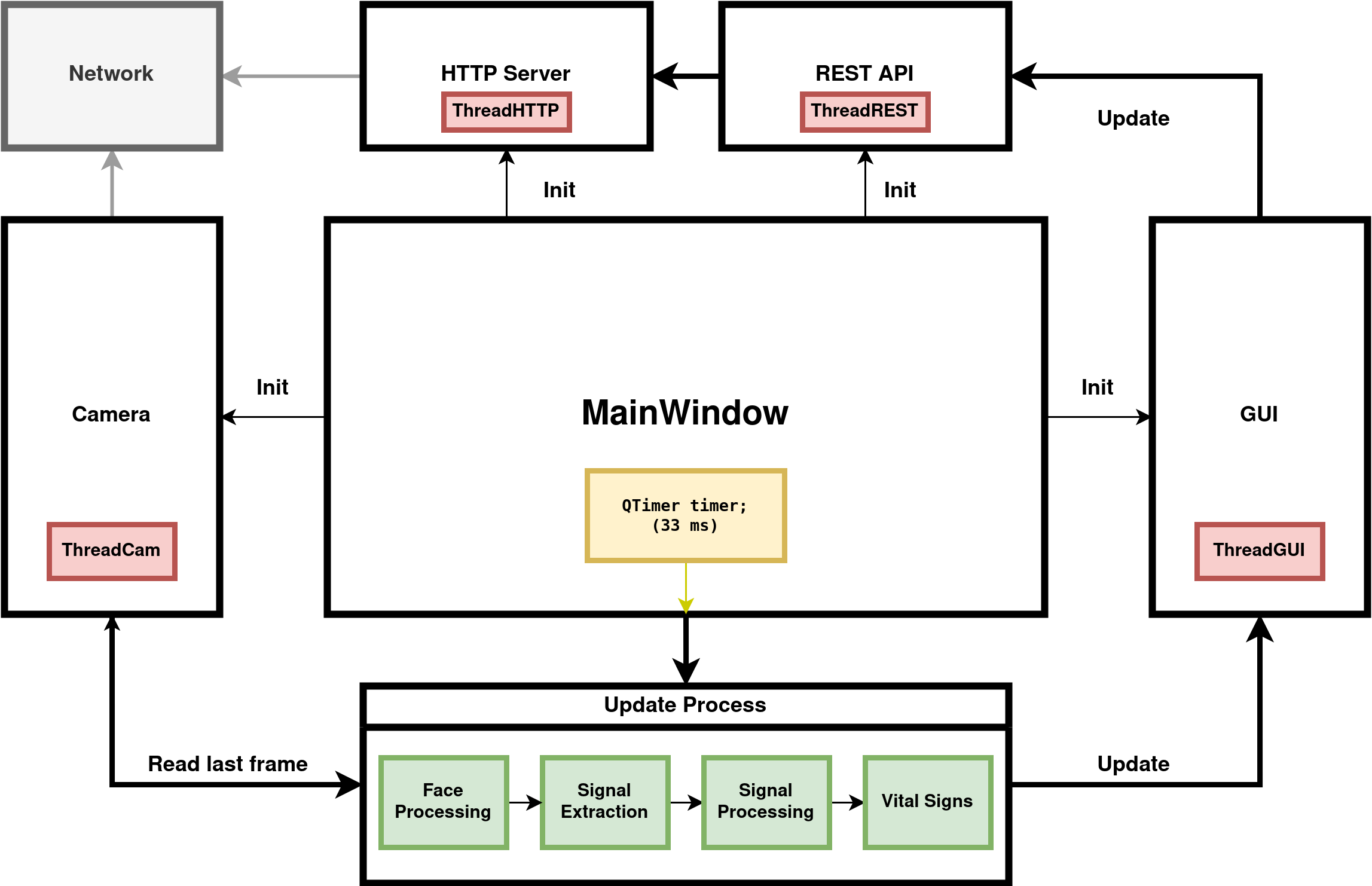}
\end{center}
\caption{Implementation design of the proposed real-time rPPG application.}
\label{fig:impl_arch_rt_app}
\end{figure}

The system architecture is organized into four primary functional blocks: the collaborative UI, video capture, real-time processing and analysis, and the network interface, as shown in Figure \ref{fig:impl_arch_rt_app}. These blocks provide the necessary infrastructure for data handling, user interaction, and network communication, while the core of the system, the optimized Face2PPG pipeline, handles the rPPG signal acquisition and analysis, allowing for the extraction of physiological signals from video data. The collaborative UI plays a key role in ensuring optimal conditions for rPPG signal extraction. It provides real-time video quality metrics and user feedback, helping to maintain consistent biosignal capture by guiding the user on appropriate positioning and environmental factors such as lighting. The video capture thread is dedicated to continuously acquiring frames from the camera. Each frame provides a sample for the rPPG analysis pipeline, and it is critical to avoid frame dropping to maintain regular sampling of the biosignal. This thread minimizes delays and compensates for camera jitter, ensuring consistent frame capture for precise signal analysis. The real-time processing and analysis block is the optimized implementation of the Face2PPG pipeline. This block operates concurrently with the other blocks and performs critical tasks such as face detection, alignment, and face normalization. The processed video data is then analyzed to extract physiological signals in real time, using optimized algorithms to support immediate and accurate analysis. The network interface operates with three different threads: two threads handle continuous video streaming via HTTP, and a third thread manages the RESTful API. The video streams include the main camera feed (cropped to the face) and a normalized pulsating face stream, both transmitted as MJPEG. The RESTful API thread responds to GET requests, providing the latest computed vital signs, including HR, RR, and oxygen saturation (SpO2). In parallel, the GUI thread continuously updates the user interface with the latest video frames and vital sign values. This ensures that the interface remains responsive and up-to-date without impacting the core data processing or network transmission tasks. By distributing these tasks across multiple threads, the system leverages multicore processors to maintain high performance and reliability. 
The multithreaded design allows for concurrent execution of tasks, \revTen{aiming to support real-time biosignal monitoring and analysis without introducing bottlenecks}, while supporting continuous user feedback and data streaming.

%
%
\subsection{Reactive-based programming design}

The application implements a reactive programming model that combines aspects of Functional Reactive Programming (FRP) and the Actor Model, making it well-suited for real-time interactive systems that require rapid responses and efficient resource management. The event-driven design of the system is central to its operation, with key computational tasks triggered dynamically by events such as face detection. This approach reflects the principles of FRP, where operations are executed only when specific conditions are met, ensuring efficient use of computational resources. Although the core pipeline is modular and sequential, its event-driven nature resembles dataflow programming, a concept related to FRP. The system reacts to inputs (such as the appearance or absence of a face) by activating the necessary processing tasks as shown in Figure \ref{fig:modular_arch_rt_app}. At the same time, regular updates are maintained through a timer-driven mechanism that coordinates the activities of the system components. The asynchronous architecture follows the Actor Model, with independent threads handling specific tasks such as video capture, processing, GUI, and network operations. These threads run concurrently and are managed by a central event loop. When possible, threads are distributed across multiple cores of the processor, enhancing parallelism. The main thread synchronizes system operations by aggregating updates from these threads at regular intervals. This design ensures that no thread blocks another or causes unnecessary waiting, even when data exchange between threads is required. As a result, the system maintains continuous, efficient operation without interruption. This hybrid model blends the declarative nature of FRP with the concurrent, message-passing framework of the Actor Model. By combining these programming paradigms, the system achieves robust real-time performance, handling both event-triggered reactions and continuous updates seamlessly within a multithreaded, multi-core environment.

The system is coordinated by a central process called \textbf{MainWindow}, which manages timing and controls the execution of pipeline processes, as shown in Figure \ref{fig:impl_arch_rt_app}. The \textit{MainWindow} initializes system components like the camera component and GUI, while also managing operational data such as the frame counter, biosignal values, and timestamps. To maintain a consistent frame rate of 30 frames per second, the \textit{MainWindow} uses the \textbf{QTimer} class from the Qt framework to trigger the main loop every 33 milliseconds. The \textit{QTimer} emits a timeout() signal at regular intervals, which activates the main loop to process frames, update the GUI, and manage other real-time tasks. In a multithreaded environment, \textit{MainWindow} interacts with other threads handling image capture, face detection, and signal processing through their event loops. This design keeps real-time performance steady while minimizing system overhead. By placing components in separate threads, the system avoids blocking the main thread during camera and GUI operations. This design ensures that image capture and GUI updates occur concurrently with other processing tasks, maintaining smooth, uninterrupted operation.  The \textit{UpdateProcessFrame} function in the \textit{MainWindow} process manages real-time processing and interface updates. It is triggered every 33 milliseconds by the \revLang{timer} in \textit{MainWindow} to maintain a consistent frame rate of 30 frames per second. The function retrieves the last available frame from the camera thread and calls the \textit{detectFacesRGB} method from the \textit{Process} class to detect faces and extract biosignals. After processing, the function updates the \textit{DataContainer} with the relevant data, including RGB frames, frame number, facial landmarks, face activity status, and calculated vital signs. A global timer tracks processing time, and the frame counter records the number of frames processed. If a face is not detected in the current frame, a counter is incremented. The system initiates a reset after 60 consecutive missed detections. Once processing is complete, the \textit{newRGBframe} flag is reset to indicate that the current frame has been fully processed and the system is ready to receive the next frame for further processing. Finally, the \textit{UpdateProcessFrame} function provides the latest data, which is then used by the \textit{MainWindow} to update the GUI and increment the frame counter. The GUI runs in a separate thread, continuously rendering the processed data for real-time display. Simultaneously, to support real-time monitoring and integration with other applications, the system includes a RESTful API and an HTTP server, both operating in their own thread. This API exposes an endpoint that accepts HTTP requests, allowing external clients to retrieve the most recent frame and computed HR, SpO2, and RR values. Upon receiving a request, the service accesses the latest data from the \textit{DataContainer} and responds with the current frame and vital sign values in a structured format.

This multithreaded design allows the system to handle incoming network requests, transmit real-time data, and update the GUI concurrently without interference with the main processing tasks. 
\revTen{The modular and thread-based architecture is designed so that camera and GUI operations, along with network services, do not interrupt the core processing tasks.} Efficient computational algorithms, optimized for hardware constraints, enable continuous biosignal extraction and vital sign calculation while ensuring effective resource management through multiprocessing and parallelization.

%
%
%
%

\subsection{Real-time implementation of the network software components}
\label{subsec:network}

Following the real-time architecture described earlier, the network components are designed to run concurrently with other system processes, ensuring consistent data transmission and integration with external applications.


\textbf{The RESTful API} is initialized in the \textit{MainWindow} constructor, where a dedicated thread is created to run the \textit{initializeREST} function. This function configures the REST service using the \textit{Restbed} library\footnote{https://github.com/Corvusoft/restbed}, which handles HTTP requests asynchronously within the dedicated thread. The service operates on port 8080, and the resource endpoint \texttt{/vhealth} is used to provide vital sign data. When a client makes a request to \revTen{\url{http://127.0.0.1:8080/vhealth}}, the system responds with seven formatted values as follows:

\begin{verbatim}
%05.1f,%05.1f,%05.1f,%d,%d,%08.1f,%08.1f
\end{verbatim}

\noindent These values correspond to the following variables:
\begin{itemize}
\item \texttt{g\_hr}: Heart Rate
\item \texttt{g\_br}: Breathing Rate
\item \texttt{g\_O2}: Oxygen Saturation (SpO2)
\item \texttt{g\_seeuser}: Indicator of whether a user is present
\item \texttt{g\_stable}: Indicator of whether the values are considered reliable
\item \texttt{g\_hr\_graph}: Latest value of the PPG signal for heart rate visualization
\item \texttt{g\_br\_graph}: Latest value of the respiratory signal for visualization
\end{itemize}

\revTen{The HR, RR and SpO2 values returned by the RESTful API are the same values that are displayed in the GUI, computed by the post-processing of the pipeline, and the \texttt{g\_stable} indicator allows client applications to discard the values obtained before the signal is stable.}

By separating the network operations into their own thread, the system ensures that the RESTful API does not block core processing tasks, maintaining real-time performance. The stateless nature of the RESTful API allows for easy integration with external applications through standardized HTTP communication, providing continuous access to vital sign data without disrupting the ongoing processing within the system.


\textbf{The HTTP server} is initialized in the \textit{MainWindow} constructor by invoking the \textit{startHttpSrvrThread} function, which takes parameters such as the port number, network interface, verbosity level, and URL prefix. The server operates in its own thread, ensuring that network operations remain separate from the main application and do not interfere with real-time processing. \textbf{Data transmission} is managed within the \textit{UpdateProcessFrame} function, which is executed every 33 milliseconds to maintain a consistent frame rate of 30 FPS. During each cycle, the latest frame and vital sign values are prepared for transmission.
\revTen{The network interface} operates in two parallel modes:

\begin{enumerate}
\item \textbf{Request-Response Mode}: \revTen{The RESTful API} listens for incoming HTTP requests on port 8080 (\texttt{/vhealth}). Upon receiving a request, it responds with the latest numerical vital sign data (heart rate, breathing rate, oxygen saturation, and related status flags). Video data is not streamed in this mode.

\item \textbf{Continuous Streaming Mode}: Simultaneously, the \revTen{HTTP} server continuously streams two video feeds without requiring explicit requests from the client:
\begin{itemize}
    \item \revTen{\texttt{http://127.0.0.1:8081/mainstream}} – Streams the cropped main camera feed as an MJPEG stream on port 8081.
    \item \revTen{\texttt{http://127.0.0.1:8081/pulsestream}} – Streams the normalized pulsating face data as an MJPEG stream on port 8081.
\end{itemize}
\end{enumerate}

In the streaming mode, MJPEG is used for video compression. This format compresses each frame independently, eliminating temporal dependencies, meaning that each frame is compressed and transmitted without needing information from adjacent frames. This approach prevents temporal artifacts and ensures that the loss of one frame does not affect the quality of subsequent frames, 
\revTen{making it suitable for real-time streaming and physiological assessment purposes}.

Both modes \revLang{(request-response and continuous streaming)} run concurrently \revTen{in the two servers}, supported by asynchronous and non-blocking I/O operations. This enables the \revTen{network interface} to handle numerical data requests while continuously pushing video streams to connected clients in real time. The separate threading ensures that network operations do not disrupt the core processing pipeline, preserving the real-time performance of the system. This architecture supports multiple simultaneous client connections, allowing for real-time monitoring across various devices.
\revTen{The performance of the network interface, including MJPEG encoding, API response time and operation with up to eight stream clients, is also evaluated during the live operation of the complete system.}

\revTen{In the publicly released code, indicated in the availability statement, both servers are implemented in a network module of our own, based on POSIX sockets, for two reasons: Restbed is distributed under the GNU Affero General Public License or a commercial license, which is not compatible with the license of the repository, and a module without external network libraries simplifies building the application on other platforms. The module keeps the same ports, endpoints and data formats, so client applications work with both implementations, and its ports and listening address can be configured to avoid conflicts with other services. Each server runs in its own thread and, like the implementation described above, it can serve clients in the local network. Security features such as authentication and encryption are outside the scope of this work.}


%
%
%
%

\section{Performance evaluation of the real-time camera-based rPPG system}

\textcolor{black}{This section outlines the experimental methodology and evaluation strategy used to assess the proposed real-time rPPG system. We describe the setup of two system configurations, detail the datasets used, and report both quantitative metrics and processing speed\revTen{, as well as the operation of the complete system with a live camera}. This structured evaluation enables us to analyze the trade-offs between computational efficiency and physiological signal accuracy across real-time and high-performance environments.}

We evaluated the performance of the real-time camera-based rPPG system in two configurations. The first operates without computational constraints, suitable for high-performance computing environments such as cloud servers or workstations, using heavy deep learning models and advanced signal processing algorithms for high-accuracy biosignal extraction. 
The second configuration is designed for \revTen{end-user equipment with limited computational resources}, with faster components and algorithmic adjustments to ensure real-time processing and efficient resource usage. Performance metrics and computational efficiency are analyzed to evaluate the balance between accuracy and real-time operation.

\subsection{Experimental setup and configurations}
\label{subsec:setup}


The experimental setup for the real-time rPPG prototype consists mainly of two configurations derived from the Face2PPG pipeline \cite{Face2PPGPipeline2022}: Face2PPG-RT, implemented in C++14 for real-time processing on \revTen{end-user equipment}, and Face2PPG-Server, implemented in Python 3.8 for high-accuracy, resource-intensive tasks. The evaluation compares their performance, focusing on computational efficiency and signal accuracy under different operational constraints. Table \ref{tab:RT_Configurations} details the modules and methods across these configurations.

\begin{table*}[ht!]
\def\arraystretch{1.2}%
\setlength{\tabcolsep}{0.8em}

\begin{center}
  \caption{\revLang{Configuration setups and modules for real-time and cloud-based rPPG systems.} \revTen{In the C++ application used for the speed evaluation, the face normalization uses a mesh of 122 triangles, and the spectral analysis of both configurations uses the FFT with a Hann window.}}
  \label{tab:RT_Configurations}
  \scalebox{0.86}{
  \begin{tabular}{lllll} 
  \toprule

    &  Face2PPG-RT
    &  Face2PPG-RT
    &  Face2PPG-Server
    &  Face2PPG-Server \\

       Module
    &  \small \textit{Config. 1}
    &  \small \textit{Config. 2}
    &  \small \textit{Normalized}
    &  \small \textit{Multiregion }  \\

\midrule

        Face Detection &  Dlib HOG & OpenCV YuNet & OpenCV SSD & OpenCV SSD  \\

        Face Alignment &  Dlib ERT GTX & Dlib ERT GTX & DAN & DAN  \\

        Face Normalization &  \makecell[tl]{Mesh \scriptsize(131 \revTen{triangles})}  &  \makecell[tl]{Mesh \scriptsize (131 \revTen{triangles})} &  \makecell[tl]{Mesh \scriptsize (131 \revTen{triangles})} &  \makecell[tl]{Mesh \scriptsize (131 \revTen{triangles})}  \\

        Skin Segmentation &  \makecell[tl]{Geometrical \\ (Patches)} & \makecell[tl]{Geometrical \\  (Patches)} & \makecell[tl]{Geometrical \\ (Whole Face)} & \makecell[tl]{Geometrical \\ (Best Regions)}  \\

        \revTen{RGB} to BVP &  \makecell[tl]{Lab} & \makecell[tl]{Lab} & \makecell[tl]{POS} & \makecell[tl]{POS}  \\

        Filtering &  \makecell[tl]{FIR BPF \\ \small (Lanczos win.)} & \makecell[tl]{FIR BPF \\ \small (Lanczos win.)} & \makecell[tl]{FIR BPF \\ \small (Kaiser win.)} & \makecell[tl]{FIR BPF \\ \small (Kaiser win.)}  \\

        Spectral Analysis &  \makecell[tl]{FFT \\ \small (Hamming win.)} & \makecell[tl]{Welch \\ \small (Hamming win.)} & \makecell[tl]{Welch \\ \small (Hamming win.)} & \makecell[tl]{Welch \\ \small (Hamming win.)}  \\

        Language &  C++ & C++ & Python & Python  \\

\bottomrule
\end{tabular}}
\end{center}
\end{table*}

Face2PPG-RT is designed for real-time performance in constrained computational platforms such as embedded systems or mobile devices. This configuration is evaluated in two setups: \textit{Configuration 1} uses the Dlib HOG-based face detector and basic FFT for frequency analysis. \textcolor{black}{In this FFT-based approach, a \revTen{10-second input signal (approximately 200, 250 or 300 samples, depending on the frame rate of the dataset)} is first \revTen{normalized and multiplied by} a Hamming window \revTen{to reduce spectral leakage}. Then, it is zero-padded to \revTen{2048 points} before computing the frequency spectrum using the DFT}\revTen{, which provides a denser sampling of the spectrum without increasing its frequency resolution, determined by the 10-second duration of the signal}\revTen{;} and \textit{Configuration 2} uses the faster OpenCV YuNet CNN \cite{Wu2023yunet} face detector and Welch’s method for spectral analysis to improve signal robustness while preserving real-time processing capabilities; \textcolor{black}{here, a \revTen{10-second} input signal is processed by dividing it into 256-sample segments with an overlap of 200 samples (approximately 78\%), applying a Hamming window to each segment, and averaging the power spectral densities derived from \revTen{8192}-point FFTs of these segments.} \revTen{When the window contains fewer than 256 samples, as in COHFACE (200 samples) and LGI-PPGI (250 samples), the segment length is set to the window length. With these parameters, each 10-second window contains a single segment, which covers the whole window in COHFACE and LGI-PPGI and its first 256 samples in UBFC-RPPG, so no averaging of periodograms takes place in our evaluation.} Both configurations use the ERT-GTX model \cite{AlvarezBordallo2021FaceAlignment} in Dlib library (version 19.23) for face alignment, geometric skin segmentation focusing on cheeks and forehead, and FIR filtering with a Lanczos window. The system, compiled using g++ 7.5.0, incorporates a network module, camera capture, and GUI for full operation. 
However, the evaluation focuses on the core video processing and signal analysis components\revTen{, which are common to both setups}.
\revTen{The HR accuracy of both Face2PPG-RT configurations reported in Table~\ref{tab:PipelinesEvaluationPPG_RT} was evaluated offline with a Python implementation of the real-time processing chain, which processes the pre-recorded dataset videos frame by frame, whereas the processing times reported in Table~\ref{tab:SpeedPerformanceRT} were measured with the C++ application. In the accuracy evaluation, the two configurations differ in the face detector (Dlib HOG or YuNet) and in the spectral analysis method (FFT or Welch's method). In the C++ application, the spectral analysis is computed with the FFT-based approach in both configurations, so the processing times of the two configurations in Table~\ref{tab:SpeedPerformanceRT} differ mainly in the face detection module.} The performance of Face2PPG-RT is enhanced by leveraging C/C++ software libraries such as \textit{Lapack} and \textit{BLAS} for efficient linear algebra operations, \textit{OpenMP} for parallelization across multiple processor cores, and \textit{FFTW3} for fast and optimized discrete Fourier transform (DFT) calculations in frequency-domain analysis. These libraries enhance computational efficiency, enabling the system to maintain real-time performance in resource-constrained environments. Face2PPG-Server prioritizes accuracy by using more computationally demanding algorithms. This configuration is tested with two pipelines: \textit{Normalized Pipeline} uses a deep learning-based Single Shot Multibox Detection (SSD) face detector, along with the Deep Alignment Network (DAN) for face alignment \cite{DANKowalskiNT17}, and processes the entire normalized facial skin region for signal extraction. \textit{Multiregion Pipeline} also uses SSD and DAN but dynamically selects the optimal skin regions for signal extraction based on quality metrics across multiple facial regions. 
\textcolor{black}{For spectral analysis, both Face2PPG-Server pipelines use Welch's method with parameters identical to those described for the Face2PPG-RT Configuration 2 (i.e., 256-sample segments with 200-sample overlap \revTen{and} Hamming windowing\revTen{, applied to 10-second windows}).}

Face2PPG-RT was tested on a laptop with an Intel\textsuperscript{\tiny\textregistered} Core i7-6700HQ CPU (\revTen{4 physical cores at 2.60 GHz, 8 hardware threads}) with Intel\textsuperscript{\tiny\textregistered} HD Graphics 530 and \revTen{16GB} of RAM, running Linux Ubuntu 18.04.6. \revTen{Face2PPG-RT supports the use of a GPU for the face detection, which is performed with the DNN module of OpenCV, but all its modules were run on the CPU in this work, without GPU acceleration, so the reported processing times also apply to devices without a GPU.} In contrast, Face2PPG-Server was evaluated on a high-performance workstation equipped with an AMD\textsuperscript{\tiny\textregistered} Ryzen 3700X 8-core processor, 64GB of RAM, and two NVIDIA GeForce\textsuperscript{\tiny\textregistered} RTX 2080 GPUs. This system can handle the heavier computational loads required by deep learning modules and complex signal processing tasks. The use of two different hardware platforms is justified by the different goals of each configuration. Face2PPG-RT was tested on mid-range hardware to simulate performance in end-user equipment, while Face2PPG-Server was tested on a high-performance system to evaluate the maximum potential of the algorithms without real-time constraints. 
\revTen{Since the configurations differ in hardware, programming language and algorithms, their processing times are reported as separate platform benchmarks rather than as a direct speed comparison}, highlighting the trade-offs between computational efficiency and accuracy in each operational context.


\subsection{Benchmark datasets and \revLang{evaluation metrics}}
\label{subsec:benchmark}

\revBoth{We evaluate the HR estimation of the rPPG system implementations using three publicly available datasets: LGI-PPGI \cite{LGIMethod2018}, COHFACE \cite{COHFACE2017}, and UBFC-RPPG \cite{UBFCDatabase2019}, the latter divided into two subsets (UBFC1 and UBFC2).} These datasets consist of RGB video recordings under different conditions, with reference physiological data collected using 
medical-grade fingertip pulse oximeters. This approach allows us to compare the extracted signals with \revTen{the reference contact measurements}. While our system is designed for real-time rPPG monitoring using live camera input, we simulate real-time conditions by processing pre-recorded videos, which provides consistent frame capture. In real-time applications, frame capture may not always be as regular, but we have implemented strategies, such as recalculating the frame rate for \revTen{the frequency analysis} \cite{Nguyen2024EvaluationOV}, to handle irregularities.
\revTen{The complete system was also evaluated during live operation, with a live camera and a reference pulse oximeter.} Rolling shutter effects occur when the camera sensor captures the image line by line, and movement during exposure causes different parts of the image to be captured at slightly different times, resulting in distortion. However, in these datasets and typical rPPG scenarios where subjects remain steady in front of the camera, notable rolling shutter effects are not expected to significantly impact performance.

COHFACE contains 160 videos of 40 subjects (12 females, 28 males, average age 35.6 years) recorded at 20 Hz and 640x480 resolution using a \textit{Logitech HD C525} webcam. Ground truth heart rate and breathing rate were measured using a fingertip pulse oximeter and respiration belt from \textit{Thought Technology}. Videos were recorded under natural and studio lighting.
LGI-PPGI includes 24 videos of 6 subjects (1 female, 5 males) aged 25-42, recorded at 25 Hz and 640x480 resolution using a \textit{Logitech HD C270} webcam. Scenarios include \revLang{``resting,'' ``rotation,'' ``talking,'' and ``gym,''} each introducing unique challenges for rPPG extraction. Heart rate was recorded using an FDA-approved \textit{pulox CMS50E} pulse oximeter at 60 Hz. UBFC-RPPG is divided into UBFC1 (8 videos) and UBFC2 (42 videos). UBFC1 captures subjects sitting still, while UBFC2 involves participants performing a mathematical task to induce blood volume pulse variations. Videos were recorded at 30 Hz and 640x480 resolution using a \textit{Logitech C920 HD Pro} webcam. Ground truth heart rate data was captured with a \textit{pulox CMS50E} pulse oximeter at 60 Hz.
\revBoth{In addition, PURE database \cite{PUREDatabase2014} is used for the preliminary evaluation of SpO2, since its reference pulse oximeter also provides SpO2 values. PURE comprises 60 videos from 10 subjects (8 males, 2 females), recorded at 30 Hz and 640x480 resolution using an \textit{eco274CVGE} camera in six scenarios with controlled head movements, and the reference data were collected with a \textit{pulox CMS50E} pulse oximeter at 60 Hz.}

\revBoth{The evaluation focuses on HR estimation, since all these datasets provide reference PPG or HR signals. Although the real-time system also estimates RR and SpO2, the evaluation of these outputs is limited in this work. Respiratory references are available only in some datasets, such as the respiration belt signals of COHFACE, which we use for a preliminary evaluation of RR. In the case of SpO2, the public rPPG datasets include healthy participants without induced hypoxemia, whose SpO2 values remain within the narrow normal range of approximately 95--100\% \cite{Mathew2023RemoteOxygen}. Under these conditions, an estimator that simply predicts a nearly constant value, for instance 98\%, would obtain a very low error without tracking any physiological change, and error metrics such as MAE or RMSE become less meaningful \cite{Mathew2023RemoteOxygen}. A meaningful validation of SpO2 requires reference measurements over a wide range of saturation values, which are typically obtained in controlled desaturation studies with induced hypoxemia, covering saturation values from 70\% to 100\% and using arterial blood samples analyzed by CO-oximetry as reference \cite{FDA2013PulseOximeters,ISO80601-2-61}. Recent camera-based SpO2 studies have followed this approach to calibrate and validate their methods \cite{vanGastel2022CalibratedSpO2,Hoffman2022SmartphoneOximetry}. Such an experiment requires a dedicated cohort and clinical protocol, and it is outside the scope of this work. Therefore, we only performed a preliminary evaluation of SpO2 on PURE, focused on whether the estimates follow the differences in SpO2 between participants.}

\textcolor{black}{To validate the performance of our system, each video from these datasets was processed. The heart rate estimations generated by our rPPG pipeline, \revTen{computed over a 10-second sliding window with updates every second, following the evaluation protocol of \cite{Face2PPGPipeline2022}}, were then temporally synchronized with the reference HR time series provided with each dataset. For each corresponding segment, the estimated HR was compared against the ground-truth HR to compute the evaluation metrics.}
\revTen{The videos were processed at their original frame rate, without resampling, and the duration of the windows was held constant, so each 10-second window contains approximately 200, 250 or 300 samples in COHFACE, LGI-PPGI and UBFC-RPPG, respectively. The HR was estimated every second, with windows centered between 5 and 50~s in COHFACE, between 10 and 55~s in LGI-PPGI, and between 5~s and 6~s before the end of each video in UBFC-RPPG. The reference HR values were computed from the reference PPG signals, resampled to the frame rate of the videos of each dataset, over the same 10-second windows with 1-second steps, and each HR estimate was compared with the reference value closest in time. Before computing the metrics, the estimated HR series of each video was smoothed with a three-sample median filter and, to compensate for its delay with respect to the reference HR series, which is caused by technical and physiological factors such as the filtering and the different blood perfusion of the face and the finger \cite{Face2PPGPipeline2022}, advanced by one second in UBFC-RPPG and by two seconds in COHFACE.}

\revBoth{We use two metrics: the Mean Absolute Error (MAE) of the HR estimates, expressed in beats per minute (bpm), and the Pearson Correlation Coefficient (PCC), which is dimensionless.} Since the blood perfusion waveforms differ between facial and fingertip regions \cite{Allen2004PPGsBody}, we use PCC between heart rate envelopes as a more appropriate metric for evaluating performance, rather than direct waveform comparisons.

\subsection{Comparative analysis of real-time and offline rPPG system configurations}

In this comparative analysis, we assess the performance of the previously depicted configurations in terms of accuracy in computing heart rate using the datasets introduced in the benchmark subsection. Moreover, we measure the speed performance of each configuration.


\subsubsection{Accuracy performance in vital signs measurement}

To assess the agreement between the rPPG systems and medical-grade \revTen{contact PPG reference devices}, we replicate the experiments conducted in \cite{Face2PPGPipeline2022}. The evaluation is conducted using the LGI-PPGI database, which includes multiple scenarios that simulate real-world conditions, such as video conferencing, gym environments, and outdoor settings with varying lighting conditions. These scenarios include both static and dynamic head movements. Additionally, we evaluated the system using the COHFACE database, which presents challenges such as heavily compressed videos and uncontrolled lighting with shadows. The UBFC-RPPG database (UBFC1 and UBFC2) is also included, providing high-quality videos recorded under controlled conditions, simulating a medical premises environment. By leveraging these datasets, we aim to evaluate the accuracy of the real-time rPPG system across different operational contexts, reflecting its performance in both controlled and uncontrolled conditions. The results are presented in Table \ref{tab:PipelinesEvaluationPPG_RT}. 
\textcolor{black}{The performance metrics reported in Table \ref{tab:PipelinesEvaluationPPG_RT} \revBoth{(MAE and PCC)} were derived by first calculating these values for each individual video sequence within a given dataset. The table presents the mean of these per-video results \revBoth{for PCC}, and the mean ± standard deviation for MAE, aggregated per dataset.}

\begin{table*}[tp]
\def\arraystretch{1.0}
\setlength{\tabcolsep}{0.8em} 
 \begin{center}
  \caption{\revBoth{Heart rate (HR) estimation} error comparison between the four different configurations of the proposed rPPG system \revBoth{on three databases, with UBFC-RPPG divided into the UBFC1 and UBFC2 subsets}. \revLang{The table includes} two configurations of the real-time system and two configurations for \revLang{high-performance} computers without any constraint. \revTen{MAE and SD are expressed in beats per minute (bpm); PCC is dimensionless.}}
  \label{tab:PipelinesEvaluationPPG_RT}
  \scalebox{0.9}{%
  \begin{tabular}{lcccccccc} 
  \toprule

       \multicolumn{1}{c}{} 
    &  \multicolumn{2}{c}{LGI-PPGI} 
    &  \multicolumn{2}{c}{COHFACE} 
    &  \multicolumn{2}{c}{UBFC1}
    &  \multicolumn{2}{c}{UBFC2} \\

\cmidrule(lr){2-3}
\cmidrule(lr){4-5}
\cmidrule(lr){6-7}
\cmidrule(lr){8-9}

       \multicolumn{1}{l}{Pipeline} 
    &  \multicolumn{1}{c}{\makecell{MAE ± SD \\ \revTen{(bpm)}}}
    &  \multicolumn{1}{c}{PCC}
    &  \multicolumn{1}{c}{\makecell{MAE ± SD \\ \revTen{(bpm)}}}
    &  \multicolumn{1}{c}{PCC} 
    &  \multicolumn{1}{c}{\makecell{MAE ± SD \\ \revTen{(bpm)}}}
    &  \multicolumn{1}{c}{PCC} 
    &  \multicolumn{1}{c}{\makecell{MAE ± SD \\ \revTen{(bpm)}}}
    &  \multicolumn{1}{c}{PCC} \\

\midrule

        \makecell[l]{Face2PPG-Server \\ \small \textit{Normalized}}   & 8.7 ± 8.4 & 0.50 & 9.4 ± 4.8 & 0.10 & 1.2 ± 0.4 & 0.94 & 1.4 ± 1.5 & 0.95 \\

        \addlinespace

        \makecell[l]{Face2PPG-Server \\ \small \textit{Multiregion}}   & 4.5 ± 3.3 & 0.57 & 8.0 ± 4.4 & 0.06 & 0.9 ± 0.4 & 0.96 & 0.9 ± 0.9 & 0.98 \\      

        \addlinespace

        \makecell[l]{Face2PPG-RT  \\ \small \textit{Configuration 1}}   &  6.4 ± 6.8 & 0.45 & \revTen{10.9 ± 5.6} & -0.04 &  \revTen{1.7 ± 0.6} & 0.80 &   \revTen{3.8 ± 4.7} & 0.72 \\

        \addlinespace

        \makecell[l]{Face2PPG-RT  \\ \small \textit{Configuration 2}}   &  5.9 ± 8.0 & 0.49 & \revTen{11.4 ± 7.3} & -0.01 &  \revTen{2.1 ± 1.1} & 0.83 &  \revTen{6.0 ± 6.3} & 0.54  \\

\bottomrule
\end{tabular}}
\end{center}
\end{table*}

\revTen{The table shows that the accuracy of the real-time setups, compared with the server setups, depends on the dataset.} Among the server configurations, using multiple regions clearly yields better outcomes. \revTen{As for real-time setups, \textit{RT Config. 2}, which utilizes Welch's frequency estimation method, obtains a lower MAE than \textit{RT Config. 1} on LGI-PPGI (5.9 vs. 6.4 bpm), but a higher MAE on COHFACE (11.4 vs. 10.9 bpm), on UBFC1 (2.1 vs. 1.7 bpm) and, especially, on UBFC2 (6.0 vs. 3.8 bpm, with a PCC of 0.54 vs. 0.72). On LGI-PPGI, both real-time configurations obtain a lower MAE than the server setup that uses simple face normalization, whereas on the UBFC subsets, recorded under controlled conditions, the server setups are clearly more accurate, especially on UBFC2, where the multiregion setup obtains an MAE of 0.9 bpm and a PCC of 0.98. On COHFACE, all the configurations obtain high errors and weak correlations, with values close to zero or slightly negative for the real-time configurations. These videos are heavily compressed and recorded under uncontrolled lighting with shadows, conditions that degrade the rPPG signal \cite{McDuff2017VideoCodecs}. Regarding the spectral analysis, which differs between the two real-time configurations together with the face detector, the HR estimates obtained with Welch's method fluctuate more between consecutive windows than those obtained with the FFT on the datasets where Configuration 2 is less accurate. On UBFC2, for example, the mean absolute difference between consecutive estimates is 2.2~bpm for Configuration 2, compared with 1.5~bpm for Configuration 1 and 1.2~bpm for the reference HR, whereas on LGI-PPGI it is lower for Configuration 2 (1.7 vs. 1.9~bpm). This suggests that, in our evaluation, the spectral estimation of Configuration 2 is more sensitive to noise. Welch's method reduces the variance of the spectrum by averaging overlapping segments, but with the 10-second windows and the segment parameters used in our evaluation, each window contains a single segment, so no averaging takes place, and in UBFC-RPPG the segment covers only 256 of the approximately 300 samples of the window, which reduces the frequency resolution. This may explain why Welch's method did not provide a consistent improvement in our experiments. These findings show that the real-time configurations can obtain an accuracy comparable to that of the server configurations in some conditions, but not in all of them, and that their performance depends on the recording conditions of each dataset.}

\revTen{To compare the two real-time configurations, which were evaluated on the same recordings, Table~\ref{tab:StatsRT} reports additional statistics of their HR estimation, and Table~\ref{tab:ScenariosRT} reports their results for the different scenarios of LGI-PPGI and the two lighting conditions of COHFACE. All the recordings of the datasets were processed (24 in LGI-PPGI, 160 in COHFACE, 8 in UBFC1 and 42 in UBFC2), and both configurations produced an HR estimate for every analysis window, so no recording or window was excluded. In COHFACE, the PCC of seven recordings in Configuration 1 and two recordings in Configuration 2 is undefined, because the estimated or the reference HR is constant during the whole recording. These recordings were counted with a PCC of zero in Table~\ref{tab:PipelinesEvaluationPPG_RT}, and excluding them gives the same rounded values. Since LGI-PPGI and COHFACE include several recordings of each participant, the 95\% confidence intervals (CIs) were computed with a bootstrap that resamples participants with all their recordings (10,000 resamples). The two configurations were compared with the paired difference of the per-recording MAE, tested with a participant-level sign-flip permutation test, and the effect size $d_z$ was computed as the mean of the participant-level differences divided by their standard deviation. In UBFC1, the participant of each recording is not identified, so each recording was treated as a separate participant. On LGI-PPGI, COHFACE and UBFC1, the differences between the two configurations are small, their CIs include zero, and they are not statistically significant. On UBFC2, the MAE of Configuration 2 is 2.2~bpm higher than that of Configuration 1 (95\% CI: 0.8 to 3.7~bpm, $p = 0.004$, $d_z = 0.45$), and the difference remains significant after the Holm correction for the four comparisons (adjusted $p = 0.017$). Regarding the scenarios, both configurations obtain the lowest errors in the resting scenario of LGI-PPGI and the highest in the gym scenario, which includes intense motion, and their errors on COHFACE are higher under natural light than under studio lighting.}

\begin{table*}[ht!]
\color{revTenColor}
\def\arraystretch{1.2}%
\setlength{\tabcolsep}{0.4em}
\begin{center}
  \caption{\revTen{Statistical comparison of the HR estimation of the two real-time configurations, evaluated on the same recordings. Rec. (Part.): number of recordings (participants). MAE: mean of the per-recording MAE, with its 95\% confidence interval (CI) computed with a participant-level bootstrap. RMSE: mean of the per-recording RMSE. $\Delta$MAE: paired difference between Configuration 2 and Configuration 1, with its 95\% CI. $p$: participant-level sign-flip permutation test. $d_z$: effect size. MAE, RMSE and $\Delta$MAE are expressed in bpm. In UBFC1, each recording is treated as a separate participant.}}
  \label{tab:StatsRT}
  \scalebox{0.8}{%
  \begin{tabular}{lcccccccc}
  \toprule
       &
    &  \multicolumn{2}{c}{Face2PPG-RT \textit{Configuration 1}}
    &  \multicolumn{2}{c}{Face2PPG-RT \textit{Configuration 2}}
    &  \multicolumn{3}{c}{Configuration 2 $-$ Configuration 1} \\
\cmidrule(lr){3-4}
\cmidrule(lr){5-6}
\cmidrule(lr){7-9}
       Dataset & Rec. (Part.) & MAE [95\% CI] & RMSE & MAE [95\% CI] & RMSE & $\Delta$MAE [95\% CI] & $p$ & $d_z$ \\
\midrule
        LGI-PPGI & 24 (6) & 6.4 [4.3, 8.6] & 8.4 & 5.9 [3.6, 9.1] & 7.6 & $-$0.5 [$-$1.5, 0.7] & 0.44 & $-$0.34 \\
        COHFACE & 160 (40) & 10.9 [9.8, 12.0] & 13.5 & 11.4 [9.8, 13.0] & 14.2 & 0.5 [$-$0.6, 1.7] & 0.41 & 0.13 \\
        UBFC1 & 8 (8) & 1.7 [1.3, 2.1] & 2.7 & 2.1 [1.5, 2.9] & 3.3 & 0.4 [$-$0.1, 1.2] & 0.30 & 0.40 \\
        UBFC2 & 42 (42) & 3.8 [2.5, 5.4] & 6.1 & 6.0 [4.1, 8.0] & 9.0 & 2.2 [0.8, 3.7] & 0.004 & 0.45 \\
\bottomrule
\end{tabular}}
\end{center}
\end{table*}

\begin{table*}[ht!]
\color{revTenColor}
\def\arraystretch{1.2}%
\setlength{\tabcolsep}{0.8em}
\begin{center}
  \caption{\revTen{HR estimation error of the two real-time configurations for the different scenarios of LGI-PPGI and the two lighting conditions of COHFACE. Rec.: number of recordings. MAE: mean of the per-recording MAE in bpm, with its 95\% CI computed with a participant-level bootstrap. PCC: mean of the per-recording PCC (dimensionless).}}
  \label{tab:ScenariosRT}
  \scalebox{0.9}{%
  \begin{tabular}{llccccc}
  \toprule
       &
    &
    &  \multicolumn{2}{c}{Face2PPG-RT \textit{Configuration 1}}
    &  \multicolumn{2}{c}{Face2PPG-RT \textit{Configuration 2}} \\
\cmidrule(lr){4-5}
\cmidrule(lr){6-7}
       Dataset & Scenario & Rec. & MAE [95\% CI] & PCC & MAE [95\% CI] & PCC \\
\midrule
        \multirow{4}{*}{LGI-PPGI} & Resting & 6 & 2.0 [0.5, 4.6] & 0.67 & 1.7 [0.6, 3.7] & 0.66 \\
         & Talking & 6 & 4.4 [3.1, 5.9] & 0.32 & 4.9 [2.2, 8.1] & 0.42 \\
         & Rotation & 6 & 7.2 [3.8, 10.8] & 0.26 & 5.6 [3.7, 7.7] & 0.27 \\
         & Gym & 6 & 11.9 [5.1, 20.8] & 0.56 & 11.3 [3.4, 23.5] & 0.60 \\
        \addlinespace
        \multirow{2}{*}{COHFACE} & Studio lighting & 80 & 9.7 [8.6, 10.9] & 0.00 & 9.7 [8.4, 11.1] & 0.00 \\
         & Natural light & 80 & 12.1 [10.5, 13.7] & $-$0.07 & 13.0 [10.8, 15.5] & $-$0.03 \\
\bottomrule
\end{tabular}}
\end{center}
\end{table*}

\subsubsection{\revBoth{Preliminary evaluation of the respiratory rate}}
\label{subsubsec:rr}

\revBoth{In addition to HR, we performed a preliminary evaluation of the RR estimation on COHFACE, the only dataset of this work with a respiration reference. The RR was estimated from the L* channel, which the application uses for RR, computed from the average color of the normalized face extracted with the YuNet face detector (\textit{RT Config. 2}) in the HR evaluation, and with the estimator of the application, i.e., the power-weighted mean frequency of the spectrum. As for HR, the RR was computed over 10-second windows with 1-second steps, centered between 5 and 50~s, without smoothing the estimates, and the reference RR was obtained from the spectral peak of the respiration belt signal over the same windows. Since all the participants were recorded at rest, the respiratory band of this evaluation was set to 6 to 24 breaths per minute (0.1 to 0.4 Hz), which covers the RR of adults at rest: the facial and belt signals were band-pass filtered within this range with a zero-phase Butterworth filter, and the spectrum was analyzed within the same range. The MAE was $3.30 \pm 1.53$ breaths per minute, with a mean bias of 0.15 breaths per minute, and it was similar under studio and natural lighting (3.47 and 3.14 breaths per minute, respectively). The RR of each participant varied little within the recordings (median standard deviation of 1.4 breaths per minute over the windows of each recording, under both lighting conditions), so the PCC within recordings, computed as for HR, was close to zero (0.03). Between recordings, the mean reference RR ranged from 6.5 to 22.3 breaths per minute, and the mean RR estimated in each recording followed these differences, with a PCC of 0.43 (95\% CI: 0.24 to 0.58, participant bootstrap). The a* and green channels and the OMIT method \cite{Face2PPGPipeline2022} obtained similar results (MAE from 3.25 to 3.35 breaths per minute and PCC between recordings from 0.36 to 0.42), and averaging the normalized spectra of the four signals (L*, a*, green and OMIT) reduced the MAE to $3.12 \pm 1.68$ breaths per minute, with a PCC between recordings of 0.46 (95\% CI: 0.23 to 0.64). These results show that the system estimates the RR from the facial video without a systematic bias, and that its estimates follow the differences in RR between participants. However, the variation of RR in this type of dataset, in which the participants are at rest, is low, so a more comprehensive and thorough evaluation, with datasets that include changes of RR, is needed for a more rigorous assessment of the RR estimation, and it is left for future work.}

\subsubsection{\revBoth{Preliminary evaluation of the oxygen saturation}}
\label{subsubsec:spo2}

\revBoth{We also performed a preliminary evaluation of the SpO2 estimation on PURE, using the 60 recordings (10 participants and six scenarios). The average red and green values of the central facial region were extracted with the face detection, alignment and normalization of \textit{RT Config. 2} (YuNet face detector), and the SpO2 was computed with Eq.~\ref{eq:spo2} over 10-second windows with 1-second steps, as for HR and RR. The reference SpO2 of each window was the mean of the values provided by the pulse oximeter over the same window. The SpO2 of the participants was within a narrow range ($97.1 \pm 1.7$\% over the windows, with mean values between 92.3\% and 99.0\% in the different recordings), and it varied little within each recording (median standard deviation of 0.3\%). The estimated SpO2 was $97.8 \pm 0.9$\%, with an MAE of $1.29 \pm 1.01$\% and a mean bias of 0.71\%. Since the reference SpO2 hardly changed within the recordings, the PCC within recordings was close to zero ($-$0.03), whereas the mean SpO2 estimated in each recording followed the differences between recordings, with a PCC of 0.41 (95\% CI: 0.06 to 0.76, participant bootstrap), and with PCCs between 0.20 and 0.55 in the six scenarios. These results show that the SpO2 estimates of the system follow the differences in SpO2 between participants. However, the range of SpO2 in this dataset is narrow, the number of participants is small and the constants of Eq.~\ref{eq:spo2} were not calibrated, so a validation in a dedicated study with induced desaturation is needed, and it is left for future work.}

\subsubsection{Speed performance}

Table \ref{tab:SpeedPerformanceRT} illustrates the speed performance of the different modules within the Face2PPG pipelines across the four proposed configurations. These configurations are divided into two primary categories: User Equipment (targeted for real-time operation) and Cloud (for unconstrained conditions, typically found in server environments).

\begin{table*}[tp]
\def\arraystretch{1.2}%
\setlength{\tabcolsep}{0.85em}
\begin{center}
  \caption{Performance speed for individual modules in various Face2PPG pipeline configurations, as well as the overall speed per face and frame. \revTen{Times are given in ms per frame, except where indicated in microseconds (us). Total FPS is the inverse of the total processing time of the core modules, and it does not include camera capture, GUI rendering or network transmission. The Spectral Analysis row includes the computation of HR, RR and SpO2 at each update of the vital signs, although the SpO2 is computed without spectral analysis.}}
  \label{tab:SpeedPerformanceRT}

  \begin{tabular}{lcccc} 
  \toprule

    \multicolumn{1}{c}{} 
    &  \multicolumn{2}{c}{User equipment configurations}
    &  \multicolumn{2}{c}{Cloud configurations} \\

\cmidrule(lr){2-3}
\cmidrule(lr){4-5}

    \multicolumn{1}{c}{} 
    &  \makecell{RT Config. 1} 
    &  \makecell{RT Config. 2} 
    &  \makecell{Server-Normalized} 
    &  \makecell{Server-Multiregion}    \\

    \multicolumn{1}{l}{Module}
    &  \small ms/frame
    &  \small ms/frame 
    &  \small ms/frame 
    &  \small ms/frame    \\

\hline

        Face Detection & 5.48 & 1.10  &  20.35 & 20.35  \\

        Face Alignment &  4.48 & 3.74 & 82.52 & 82.52  \\

        Face Normalization &  3.67 &  4.37 & 12.48  & 12.48 \\

        Skin Segmentation & 8.23 us & 8.23 us & 95.57 us & 105.41  \\

        \revTen{RGB} to BVP &  0.462 & 0.462 & 0.0798 &  0.0798 \\

        Filtering & 1.432 us  & 1.432 us & 0.163 us & 0.162 us  \\
     
        Spectral Analysis & 5.85 us & 5.85 us & 1.02 & 1.02 \\
      
        Total &  14.11 &  9.69 & \revTen{116.55} &  221.87 \\

        \addlinespace
        Total FPS & 71  & 103 & \revTen{8.58} &  4.51 \\

\hline

\end{tabular}
\end{center}
\end{table*}

The outcomes for \textit{RT Config. 1} reveal that the processing time required for each frame is approximately 14.11 milliseconds (ms). The most significant contributor to this processing time is the \revLang{`Face Detection'} module, which consumes 5.48 ms per frame. 
\revTen{This corresponds to a processing capacity of approximately 71 FPS, above the 30 FPS targeted by the system.} In contrast, the findings for \textit{RT Config. 2} showcase a more optimized performance, with each frame taking only about 9.69 ms to process. Notably, the \revLang{`Face Detection'} time has been substantially reduced in this configuration, clocking in at 1.1 ms, as compared to \textit{RT Config. 1}. 
\revTen{Consequently, the processing capacity increases to approximately 103 FPS. These FPS values are computed as the inverse of the total processing time of the core modules, which excludes camera capture, GUI rendering and network transmission. Therefore, they represent the processing capacity of the pipeline on the evaluated laptop rather than the end-to-end frame rate of the complete system during live operation\revTen{, which is evaluated later, together with the processing times of the core modules in repeated sessions on the same laptop}.}

In server conditions without real-time constraints, more intensive processing is feasible. This is evident from the \textit{Server-Normalized} configuration, where the total processing time per frame is \revTen{116.55} ms, significantly higher than the \revTen{real-time configurations}.
\revTen{This corresponds to a processing capacity of 8.58 FPS.} Remarkably, \revLang{`Face Alignment'} contributes substantially to this processing time, requiring up to 82.52 ms.Meanwhile, the \textit{Server-Multiregion} configuration \revLang{requires additional processing}, leading to a total frame processing time of 221.87 ms. This nearly doubles the processing time compared to the \textit{Server-Normalized} configuration, resulting in a relatively lower FPS of 4.51. A key observation is the \revLang{`Skin Segmentation'} module, which, in this configuration, demands 105.41 \revTen{ms}, underscoring the complexity introduced by multi-region analysis.

Table \ref{tab:SpeedPerformanceRT} provides an in-depth exploration of the speed performance of the proposed Face2PPG pipeline across various configurations. While the \revTen{real-time} configurations are tailored for real-time applications, the cloud configurations, designed for server environments, accommodate more intensive data processing and analysis, although at a reduced FPS.
\revTen{As the real-time and server configurations were measured on different hardware, with different programming languages and algorithms, these results should be interpreted as separate platform benchmarks, and the differences in speed cannot be attributed to the proposed architecture alone. In addition, on the workstation on which the server configurations were measured, the core modules of the real-time application took 5.9~ms per frame during the live evaluation, compared with 116.55 and 221.87~ms for the server configurations, implemented in Python, so the difference in processing time remains large on the same hardware, although it still combines the effects of the programming language and of the algorithms of each configuration.}

\subsubsection{\revTen{Live operation of the complete system}}
\label{subsubsec:live}

\begin{figure}[ht!]
\begin{center}
\includegraphics*[width=0.47\textwidth]{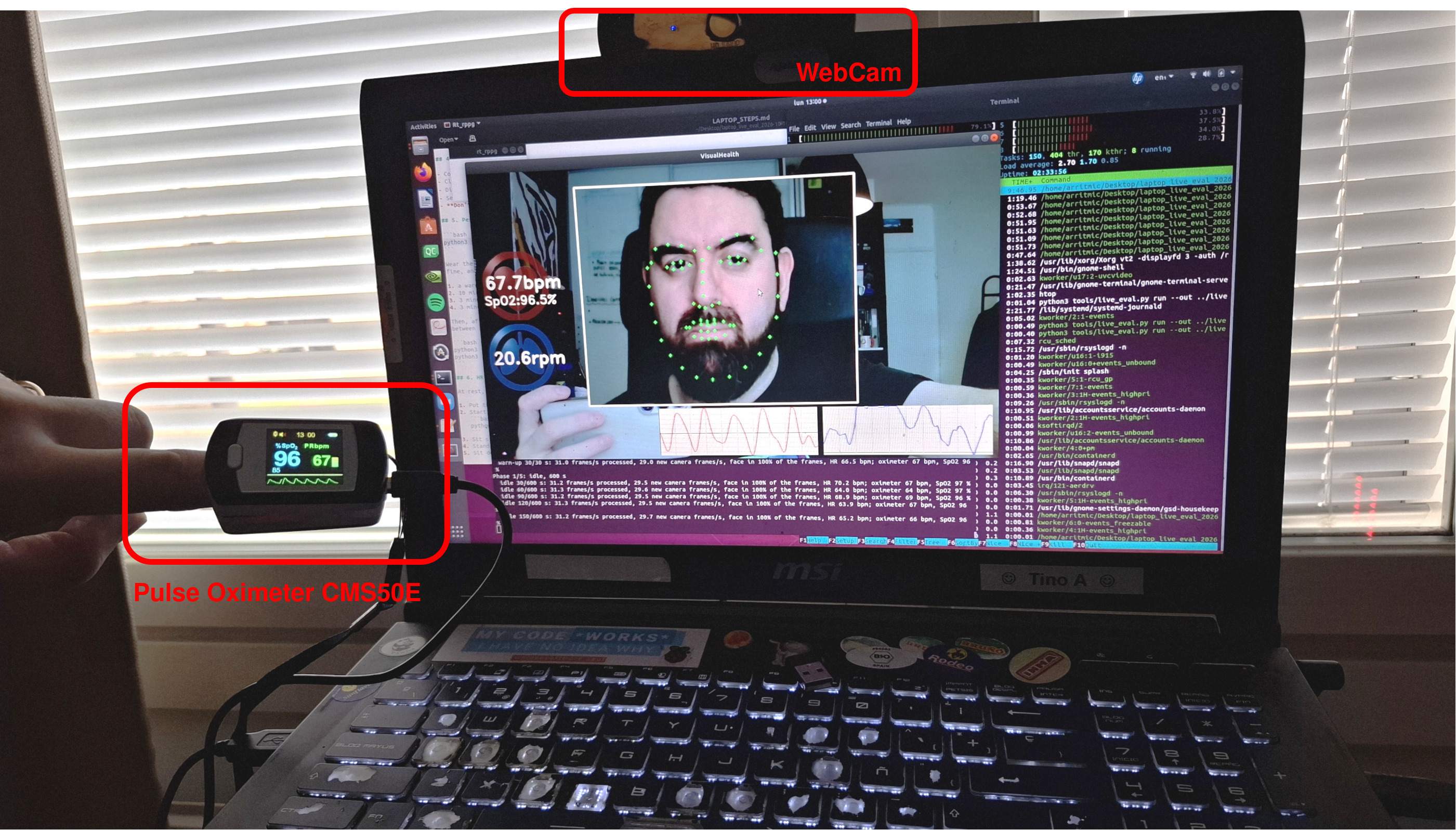}
\end{center}
\caption{\revTen{Setup of the live evaluation: the application during live operation on the laptop, in the first session, with the USB webcam on top of the screen, showing the facial landmarks, the HR, SpO2 and RR values and the pulse and respiratory signals, next to the fingertip pulse oximeter (Contec CMS50E) used as reference, connected to the same computer. The terminals behind the application window run the measurement script and a system monitor (htop), whose meters, at the top right, show the processing load distributed across the cores of the processor by the threads of the application.}}
\label{fig:live_setup}
\end{figure}

\revTen{The accuracy and speed evaluations above cover the core processing modules. To evaluate the complete system during live operation, including camera capture, GUI rendering, MJPEG encoding and the network interface, we ran the publicly released version of the application, which has the same core processing modules as in the speed evaluation, with a live camera on two computers: the laptop used for Face2PPG-RT and the workstation used for Face2PPG-Server, hereafter referred to as the desktop computer. On the laptop, the application was built with the same library versions and compiler options as in the speed evaluation, and on the desktop computer, running Linux Ubuntu 24.04, with the libraries of this distribution (Qt 5.15, OpenCV 4.6 and Dlib 19.24) and the same compiler options. The camera was a low-cost USB webcam (Trust USB Camera), chosen as an example of the consumer-grade webcams commonly used with laptops and desktop computers, providing 1280$\times$720 MJPEG images at 30~FPS, with a fixed exposure time of 33~ms and a fixed gain, which prevent the automatic exposure of the camera from lowering its frame rate under indoor lighting. The gain was adjusted to the lighting of each session, since the sessions were recorded with daylight, except the first session on the desktop computer, which was recorded in the evening with indoor lighting. One of the authors sat at rest in front of the camera during the sessions, wearing a medical-grade fingertip pulse oximeter (Contec CMS50E) connected to the same computer through USB, as shown in Figure~\ref{fig:live_setup}, and took a few deep breaths during the sessions so that the HR varied slightly. On each computer, a session of 22 minutes, after a warm-up of 30~s, consisted of five consecutive phases with increasing network load: 10 minutes without network clients, 3 minutes with requests to the RESTful API at 10 requests per second, and 3 minutes each with the same API requests plus one, four and eight clients of the MJPEG streams (from four clients on, half of the clients received each stream). Two additional sessions of 10 minutes without network clients were run on the laptop, and one on the desktop computer, to assess the variability between sessions. The clients ran on the same computer and connected through the loopback interface, so the measurements include the image encoding and the transmission by the application, but not the latency of a physical network.}

\revTen{For this evaluation, the application recorded the timestamp assigned by the camera driver (Video4Linux2) to every captured frame and, for every iteration of the main loop, the camera frame used and the time at the end of each stage (frame retrieval, processing, GUI image and network step), using the monotonic clock of the operating system. Each JPEG image sent to the stream clients carried the number of the loop iteration, so that the clients could compute the latency from the camera driver to the reception of the image. The CPU time of each thread and the memory of the application were sampled every second, and the pulse rate and SpO2 of the oximeter, provided once per second, were recorded with the same clock and paired with the HR and SpO2 displayed by the application at the same time. The pulse wave of the oximeter, provided with 60 samples per second, was also recorded, so that the beat-to-beat pulse rate could be obtained from the intervals between its peaks. The latency is measured from the timestamp of the camera driver, so it does not include the exposure and the processing inside the camera, nor the refresh of the screen. The timing log and the measurement scripts are included in the released code.}

\begin{table*}[ht!]
\color{revTenColor}
\def\arraystretch{1.15}%
\setlength{\tabcolsep}{0.6em}
\begin{center}
  \caption{\revTen{Live operation of the complete system with a live camera at 30~FPS and one participant at rest, on the laptop of the speed evaluation (Intel Core i7-6700HQ; three sessions, one of them with network load) and on a desktop computer (AMD Ryzen 7 3700X; two sessions, one of them with network load). No clients: phases without network clients. API + 8 streams: requests to the RESTful API at 10 per second together with eight clients of the MJPEG streams, on the same computer. All phases: all the phases of all the sessions, including those with the API requests alone and with one and four stream clients. Values are medians with the 99th percentile in brackets, unless otherwise indicated. Latencies are measured from the timestamp of the camera driver. Loop work: time from the start of a loop iteration to the end of its network step. Agreement: differences with a fingertip pulse oximeter, one paired value per second.}}
  \label{tab:LiveRT}
  \scalebox{0.85}{%
  \begin{tabular}{lcccccc}
  \toprule
    & \multicolumn{3}{c}{Laptop (Intel Core i7-6700HQ)} & \multicolumn{3}{c}{Desktop (AMD Ryzen 7 3700X)} \\
\cmidrule(lr){2-4}
\cmidrule(lr){5-7}
    & \makecell{No\\clients} & \makecell{API +\\8 streams} & \makecell{All\\phases} & \makecell{No\\clients} & \makecell{API +\\8 streams} & \makecell{All\\phases} \\
    Sessions & 3 & 1 & 3 & 2 & 1 & 2 \\
    Duration (s) & 1800 & 180 & 2520 & 1200 & 180 & 1920 \\
\midrule
    \multicolumn{7}{l}{\textit{Frame rate and timing}} \\
     \hspace{2mm} Loop iterations per second & 31.25 & 31.25 & 31.25 & 31.25 & 31.24 & 31.25 \\
      \hspace{2mm} New camera frames per second & 29.56 & 29.51 & 29.54 & 29.81 & 29.80 & 29.83 \\
      \hspace{2mm} Camera frames not processed (\%) & 1.5 & 1.7 & 1.6 & 0.7 & 0.7 & 0.6 \\
      \hspace{2mm} Loop interval (ms), mean $\pm$ SD & 32.0 $\pm$ 0.6 & 32.0 $\pm$ 1.4 & 32.0 $\pm$ 0.8 & 32.0 $\pm$ 0.7 & 32.0 $\pm$ 0.8 & 32.0 $\pm$ 0.7 \\
    \hspace{2mm} Loop intervals over 50~ms (number) & 3 & 0 & 3 & 4 & 2 & 6 \\
    \hspace{2mm} Loop work (ms) & 19.3 [26.3] & 24.6 [30.1] & 20.2 [28.5] & 12.6 [17.4] & 13.6 [18.1] & 13.0 [17.7] \\
    \hspace{2mm} Latency to the GUI image (ms) & 82.1 [101.0] & 84.5 [102.9] & 82.7 [101.7] & 67.1 [85.2] & 66.1 [84.8] & 67.0 [85.1] \\
    \addlinespace
    \multicolumn{7}{l}{\textit{Resources}} \\
    \hspace{2mm} CPU (\% of one core), mean & 313 & 334 & 317 & 115 & 121 & 117 \\
    \hspace{2mm} Memory (MB), maximum & 184 & 184 & 184 & 260 & 260 & 260 \\
    \addlinespace
    \multicolumn{7}{l}{\textit{Network interface}} \\
    \hspace{2mm} API response time (ms) & -- & 0.72 [4.87] & 0.73 [4.84] & -- & 0.50 [0.85] & 0.51 [0.84] \\
    \hspace{2mm} API requests (errors) & -- & 1801 (0) & 7204 (0) & -- & 1801 (0) & 7204 (0) \\
    \hspace{2mm} JPEG encoding, camera / face (ms) & -- & 2.98 / 1.00 & -- & -- & 0.70 / 0.18 & -- \\
    \hspace{2mm} Images per second per stream client & -- & 31.3 & -- & -- & 31.2 & -- \\
    \hspace{2mm} Stream data, all clients (Mbit/s) & -- & 20.7 & -- & -- & 22.5 & -- \\
    \hspace{2mm} Latency to the stream clients (ms) & -- & 89.6 [108.7] & 89.3 [108.4] & -- & 67.8 [86.5] & 68.4 [87.0] \\
    \addlinespace
    \multicolumn{7}{l}{\textit{Agreement with the pulse oximeter}} \\
    \hspace{2mm} HR MAE (bpm) & 2.0 & 1.6 & 1.8 & 2.6 & 1.9 & 2.4 \\
    \hspace{2mm} HR RMSE (bpm) & 4.2 & 5.5 & 3.9 & 5.6 & 3.5 & 5.0 \\
    \hspace{2mm} HR bias (bpm) & 0.1 & 0.7 & 0.1 & 0.3 & $-$0.2 & 0.0 \\
    \hspace{2mm} HR within 3 bpm (\%) & 82 & 95 & 85 & 77 & 83 & 80 \\
    \hspace{2mm} SpO2 MAE (\%) & 1.1 & 1.1 & 1.1 & 1.1 & 1.3 & 1.0 \\
\bottomrule
\end{tabular}}
\end{center}
\end{table*}

\revTen{Table~\ref{tab:LiveRT} summarizes the results. On both computers, the main loop ran between 31.20 and 31.25 times per second in every minute of the sessions, and the camera delivered 30.0 frames per second. Since the QTimer, set to 33~ms, triggered the main loop every 32.0~ms on average, about 5\% of the iterations processed the latest camera frame again, and 1.6\% and 0.6\% of the camera frames were replaced by the next frame before being processed on the laptop and on the desktop computer, respectively. The interval between iterations had a standard deviation of 0.8 and 0.7~ms, and only three and six intervals exceeded 50~ms (maximum of 61 and 101~ms). On the laptop, these three intervals occurred during an episode of about 40~s in the first session, in which other processes increased the CPU load of the laptop and used all its hardware threads for about 5~s. During this episode, 5 camera frames were lost and the latency increased up to 250~ms, and no other camera frames were lost during the sessions. The work of each iteration, from the retrieval of the frame to the network step, took a median of 20.2~ms on the laptop and 13.0~ms on the desktop computer (99th percentiles of 28.5 and 17.7~ms), within the 32-ms interval in all the phases, and the latency from the camera driver to the GUI image had a median of 83 and 67~ms (99th percentiles of 102 and 85~ms). This latency was dominated by the transfer and decoding of the MJPEG images (median of 47 and 37~ms from the driver timestamp to the decoded frame) and by the wait for the next loop iteration (median of 16~ms on both computers), whereas the processing and the GUI took a median of 20 and 13~ms. On the laptop, the application used 317\% of one CPU core on average (80\% in the main loop, 43\% in camera capture and MJPEG decoding, and 193\% in the worker threads of the libraries), i.e., about 40\% of its eight hardware threads, and its memory remained between 180 and 184~MB. On the desktop computer, it used 117\% of one CPU core (55\%, 15\% and 46\%, respectively), i.e., about 7\% of its 16 hardware threads, and between 255 and 260~MB of memory. The dedicated GPUs of the computers were not used.}

\revTen{The processing times of the core modules, printed by the application for every frame, were also recorded during the sessions. On the laptop, the core modules took 8.9, 8.0 and 8.1~ms per frame on average in the three sessions (face detection 1.15 to 1.18~ms, face alignment 2.84 to 2.95~ms, face normalization 3.63 to 4.44~ms and RGB to BVP 0.34~ms, with less than 0.02~ms for each of the remaining modules), close to the 9.69~ms of RT Config. 2 reported in Table~\ref{tab:SpeedPerformanceRT}, which uses the same face detector. On the desktop computer, they took 5.9~ms per frame in both sessions.}

\revTen{The network load had a small effect on the operation of the system. On each computer, the API answered all the 7,204 requests without errors, with a median response time of 0.7 and 0.5~ms on the laptop and on the desktop computer (99th percentiles of 4.8 and 0.8~ms), and the encoding of the JPEG images took a median of 3.0 and 0.6 to 0.7~ms per iteration for the camera stream and of 0.9 to 1.0 and 0.2~ms for the stream of the normalized pulsating face. With eight stream clients, with respect to the phase without clients of the same session, the median work per iteration increased by 4.6 and 0.9~ms and the CPU use by 18\% and 5\% of one core, each client received 31.2 to 31.3 images per second (20.7 and 22.5~Mbit/s in total), and the median latency from the camera driver to the reception by the clients was 90 and 68~ms, less than 1~ms after the images were ready. The phases with the API requests alone and with one and four stream clients gave intermediate values.}

\revTen{During the sessions, the HR displayed by the application differed from the pulse rate of the oximeter by an MAE of 1.8~bpm on the laptop (2,520 paired values; bias of 0.1~bpm, RMSE of 3.9~bpm; MAE of 1.7, 1.9 and 1.7~bpm in the three sessions) and of 2.4~bpm on the desktop computer (1,920 paired values; bias of 0.0~bpm, RMSE of 5.0~bpm; MAE of 2.1 and 3.1~bpm in the two sessions), with 85\% and 80\% of the values within 3~bpm and 98\% and 95\% within 10~bpm. In the second session on the desktop computer, the HR of the application deviated from the pulse rate during several short periods, which increased its MAE. The MAE did not increase with the network load (2.3 and 2.2~bpm without clients and 1.6 and 1.9~bpm with eight stream clients, in the sessions with network load). The PCC was 0.66 and 0.50, limited by the narrow range of HR of the participant (58 to 85~bpm according to the oximeter), and the 95\% limits of agreement were $-$7.5 to 7.7~bpm and $-$9.7 to 9.7~bpm, widened by short periods with larger differences, as shown in Figure~\ref{fig:live_agreement}. 

\begin{figure}[ht!]
\begin{center}
\includegraphics*[width=0.49\textwidth]{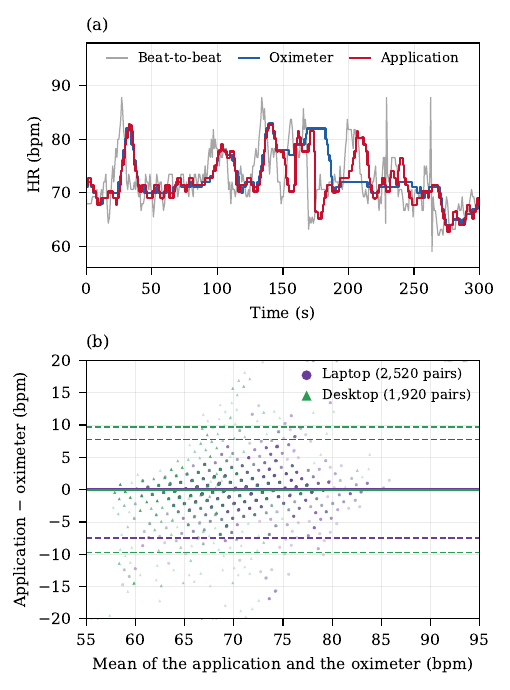}
\end{center}
\caption{\revTen{Agreement of the HR of the application with the fingertip pulse oximeter during live operation. (a) HR displayed by the application, pulse rate displayed by the oximeter and beat-to-beat pulse rate obtained from the pulse wave of the oximeter, in the first 300~s of the third session on the laptop. (b) Bland-Altman plot of the 4,440 paired values, one per second, of the five sessions, with the bias (solid lines) and the 95\% limits of agreement (dashed lines) of each computer; the 24 differences larger than 20~bpm are drawn at the border.}}
\label{fig:live_agreement}
\end{figure}

As depicted in Figure~\ref{fig:live_agreement}a, the HR of the application and the pulse rate of the oximeter followed the same changes of the pulse rate, such as the increases caused by the deep breaths. With respect to the beat-to-beat pulse rate obtained from the pulse wave of the oximeter, and estimated as the lag of maximum cross-correlation of the values sampled every second, the HR of the application followed these changes with a delay of about 6~s in all the sessions, mainly due to the 10-s window of the spectral analysis, whereas the delay of the pulse rate displayed by the oximeter was 6 to 7~s in four sessions and 12~s in the other one. In the 24 increases of the beat-to-beat pulse rate of at least 8~bpm in the five sessions, the HR of the application rose by a median of 107\% of the increase, and by at least half of it in 23 of them, although in two of them it also displayed erroneous values that exceeded the increase. The pulse rate displayed by the oximeter rose by a median of 72\% of the increase, and by less than half of it in 9 of them, since the oximeter smooths its displayed values and, at times, keeps them constant for several seconds. The SpO2 displayed by the application differed from that of the oximeter by an MAE of 1.1\% and 1.0\% (bias of 0.3\% and 0.4\%), with oximeter values between 94\% and 99\%. These results show that the complete system, including camera capture, GUI and network interface with up to eight stream clients, processed the camera frames at the frame rate of the camera during all the sessions, with a stable latency except during the episode of high load described above, both on the laptop used for the speed evaluation and on a more powerful desktop computer. However, they correspond to one participant at rest, with the network clients on the same computer.}


%
%
%
%
\section{\revBoth{Limitations}}
\label{sec:limitations}

\revBoth{The present work has several limitations. First, the HR accuracy was evaluated by processing pre-recorded videos from public datasets, which provide regular frame capture, rather than live camera streams. \revTen{The accuracy during live operation was only measured against a pulse oximeter in five sessions with one participant at rest and a narrow range of HR, so its evaluation with more participants, head motion and changes of HR remains to be done.} Second, the processing times reported in Table~\ref{tab:SpeedPerformanceRT} correspond to the core video processing and signal analysis modules\revTen{, whereas the complete system, including camera capture, GUI rendering, MJPEG encoding and the network interface, was evaluated during live operation with the network clients on the same computer, so the latency over a physical network was not measured.} In addition, the real-time and server configurations were measured on different hardware \revTen{in Table~\ref{tab:SpeedPerformanceRT}, and they differ in programming language and algorithms}, so their processing times are separate platform benchmarks rather than a controlled comparison. Third, the system was evaluated only on a laptop \revTen{and a desktop computer}, and no measurements of power consumption, energy per frame or thermal behavior were performed on embedded or low-power platforms.}

\revBoth{Fourth, the public datasets used in this work include a limited number of participants, and their demographic and skin tone diversity is limited and not fully reported, although skin tone is known to affect rPPG performance \cite{Setchfield2024TypeSkin}. Fifth, the system assumes that the user remains relatively still in front of the camera, and the accuracy is lower on the datasets that include head motion (LGI-PPGI) or heavily compressed videos recorded under uncontrolled lighting (COHFACE). Sixth, although the system estimates RR and SpO2, the quantitative evaluation focuses on HR, and HRV is not computed by the real-time system. The RR and SpO2 were only evaluated in a preliminary way, on COHFACE and PURE, respectively, with low variations of RR and a narrow range of SpO2, and the validation of SpO2 requires a dedicated study including induced desaturation. Finally, the system has not been clinically validated. The results presented in this work demonstrate an engineering prototype for real-time HR estimation, and its use in clinical or telemedicine applications would require further validation with appropriate cohorts and reference devices.}

%
%
%
%
\section{Conclusion}

Current biosignal monitoring systems often exhibit limitations in key areas such as real-time data processing, system scalability, and interoperability. To address these challenges, this work focused on the design and implementation of a real-time remote photoplethysmography (rPPG) system capable of non-contact heart rate measurement using RGB cameras, with integrated network capabilities for data transmission via a RESTful API and HTTP streaming. 
The system includes an event-driven user interface, carefully optimized for \revTen{computational efficiency}. The real-time rPPG system is structured to meet the requirements of both resource-constrained devices and high-performance systems. It follows a modular architecture, comprising distinct components for sensing, processing, communications, storage, and user interaction. These components were optimized to ensure continuous real-time operation and stable performance, even under conditions such as irregular frame rates and potential frame drops. Key optimizations include efficient algorithms for face detection, alignment, and signal processing, as well as dynamic adjustments in the processing pipeline to maintain accuracy. Additionally, real-time quality metrics, such as head movement tracking, were implemented to assess the reliability of biosignal extraction. These metrics provide feedback to the user to enhance their positioning and environmental conditions, improving the accuracy  and efficiency of the measurements. 
\revTen{Previous evaluations of rPPG methods under network and computational constraints showed that real-time challenges, including frame dropping and reduced frame rates caused by network constraints or low-light conditions, can be mitigated by recalculating frame rates for filtering and frequency-domain analysis, with a limited loss of accuracy \cite{Nguyen2024EvaluationOV}.} The results suggest that the system can operate effectively with reduced data, up to 16 times less in some scenarios, without substantially compromising rPPG measurement accuracy \cite{Nguyen2024EvaluationOV}. 
\revTen{This could reduce the computational load in embedded applications, although the power consumption of the system was not measured in this work.} These findings are consistent with studies such as Mohan et al. (2018), which showed that performance in object tracking could be maintained with significantly lower frame rates when appropriate strategies are applied \cite{Mohan2018FrameRateDropObjectDetect}. 
\revTen{Moreover, the core modules of the real-time rPPG system required between 9.69 and 14.11 ms per frame on a laptop, while the server-based configurations required between 116.55 and 221.87 ms per frame on a workstation, although these values correspond to different hardware and implementations. In terms of accuracy, the real-time configurations obtained results comparable to those of the server-based configurations on LGI-PPGI, but a lower accuracy on the UBFC subsets, and all the configurations showed a limited accuracy on the highly compressed COHFACE videos. The low processing time of its core modules highlights the potential of the system for deployment in devices with limited computational power, such as mobile or embedded systems, which should be confirmed with specific evaluations on those platforms.}
\revTen{During live operation with a camera at 30~FPS, on the laptop used for the speed evaluation and on a desktop computer, the complete system, including the GUI and the network interface with up to eight stream clients, processed the camera frames at the frame rate of the camera during all the sessions, with a median latency of 83 and 67~ms from the camera driver to the GUI image, and its HR differed from that of a fingertip pulse oximeter by an MAE of 1.8 and 2.4~bpm for one participant at rest.}
\revBoth{We evaluated the HR accuracy using three publicly available datasets}, which include varying environmental conditions such as low light, low dynamic range with strong backlighting, and highly compressed videos. These tests allowed us to assess the robustness under \revTen{challenging recording} conditions. 

\textcolor{black}{Beyond its direct applications, this work offers specific value for other researchers. The system implementation in C++, with minimal third-party dependencies, differs from many rPPG prototypes typically developed for offline analysis or in Python, and has been designed \revTen{with deployment on embedded platforms in mind}. Its modular and concurrent architecture facilitates adaptation for different experimental setups or integration into more complex clinical interfaces, allowing researchers to build upon a stable real-time foundation. The design and C++ implementation, optimized for \revTen{computational efficiency} with minimal dependencies, thus constitute a practical resource for the research community. Furthermore, the technical design of the system provides a reference for real-time biomedical signal processing, particularly in addressing challenges related to scheduling, concurrency, and latency under operational constraints. The analysis of these engineering aspects reveals important trade-offs between system responsiveness, computational load, and the fidelity of physiological signals. These insights are also relevant to fields such as embedded systems, robotics, and hybrid intelligence, contributing to the broader development of more robust and efficient context-aware computing systems.}

Our contributions aim to solidify rPPG as a viable and versatile tool within modern healthcare and human-computer interaction, offering insights into the creation and design of more reliable and accurate real-time rPPG systems. Future work will focus on further optimizing the system for a range of deployment scenarios, including increasing robustness under variable network and hardware conditions, as well as expanding its applicability in distributed healthcare environments.
\revBoth{Future work will also include a more comprehensive validation of RR using datasets with changes of RR, the validation of SpO2, with calibrated constants, in a dedicated study with induced desaturation, and the evaluation of the complete system during live operation \revTen{with more participants and conditions, over physical networks and on embedded platforms, including its power consumption}.}

\bmhead{Acknowledgements}
The research was supported by \revLang{the} University of Oulu, the 6G Flagship (369116), Profi5 HiDyn programme (326291), and Profi7 Hybrid intelligence program (352788), funded by the Research Council of Finland, and the Orion Research Foundation (Grant: Apurahat Constantino Álvarez Casado). \revTen{The authors wish to acknowledge the use of AI-assisted tools for language editing, with a focus on grammar checking and improving readability, and for coding support. In addition, the Elicit platform was used during the literature search as an AI-assisted discovery aid for candidate records. All inclusion decisions, exclusions, data extraction, synthesis, interpretation, and final writing decisions were performed by the authors.}

\section*{Declarations}

\subsection*{Funding}
The research was supported by \revLang{the} University of Oulu, the 6G Flagship (369116), Profi5 HiDyn programme (326291), and Profi7 Hybrid intelligence program (352788), funded by the Research Council of Finland, and the Orion Research Foundation (Grant: Apurahat Constantino Álvarez Casado).

\subsection*{Conflict of interest/Competing interests}
The authors declare that they have no conflict of interest.

\subsection*{Ethics approval and consent to participate}
Not applicable

\subsection*{Consent for publication}
Not applicable

\subsection*{Availability of data and materials}
\textcolor{black}{The datasets analyzed during the current study are publicly available and detailed in Section \ref{subsec:benchmark}. \revTen{The measurements of the live evaluation, without images or video, are included in the code repository indicated below.} The source code for the system described in this article is available on GitHub at \revTen{\url{https://github.com/Multimodal-Sensing-Lab/rt-rppg}}.}
\revTen{The version corresponding to the experiments reported in this article, including the C++ real-time application used for the speed evaluation, with the network module and the SpO2 computation described in this article, and the timing log, scripts and measurements of the live evaluation, together with a script that runs the application on the videos of the datasets as pre-recorded input, is release v1.0.0 (commit \texttt{643b4d0}) of this repository, archived on Zenodo with DOI \url{https://doi.org/10.5281/zenodo.23485903}.}

\subsection*{Author Contributions}
Constantino Álvarez Casado: Writing – review \& editing, Writing – original draft, Visualization, Validation, Software, Methodology, Investigation, Formal analysis, Conceptualization.  
Sasan Sharifipour: Writing – review \& editing, Methodology, Investigation, Software.  
Manuel Lage Cañellas: Writing – review \& editing, Software, Methodology, Investigation.  
Nhi Nguyen: Writing – review \& editing, Validation, Software, Investigation.  
Le Nguyen: Writing – review \& editing, Validation, Methodology, Investigation, Formal analysis.  
Miguel Bordallo López: Writing – review \& editing, Writing – original draft, Visualization, Validation, Supervision, Software, Methodology, Investigation, Formal analysis, Conceptualization, Funding acquisition.

All authors reviewed the manuscript.

\bibliography{references}

\end{document}